\documentclass[10pt,twocolumn,letterpaper]{article}

\usepackage{iccv}
\usepackage{times}
\usepackage{epsfig}
\usepackage{graphicx}
\usepackage{amsmath}
\usepackage{amssymb}

\usepackage{xcolor}
\usepackage{soul}
\usepackage{soulpos}
\usepackage{breqn}
\usepackage[accsupp]{axessibility}

\colorlet{myblue}{cyan!10}
\DeclareRobustCommand{\hlblue}[1]{{\sethlcolor{myblue}\hl{#1}}}

\usepackage{subcaption}
\newcommand\Tstrut{\rule{0pt}{2.6ex}}         
\usepackage{amsthm}
\usepackage{mathtools}
\usepackage{amssymb}
\usepackage{amsmath}
\usepackage{textcomp}
\usepackage{multirow}
\usepackage{enumitem}
\usepackage{balance}
\usepackage{color,colortbl}
\definecolor{Gray}{gray}{0.9}
\usepackage{dblfloatfix}
\usepackage{booktabs}
\everypar{\looseness=-1}
\usepackage[font=small,skip=1pt]{caption}
\setlist[itemize]{align=parleft,left=-2mm}
\makeatletter
\newcommand*{\inlineequation}[2][]{%
  \begingroup
    \refstepcounter{equation}%
    \ifx\\#1\\%
    \else
      \label{#1}%
    \fi
    \relpenalty=10000 %
    \binoppenalty=10000 %
    \ensuremath{%
      #2%
    }%
    ~\@eqnnum
  \endgroup
}
\makeatother

\usepackage{amsmath}

\usepackage[breaklinks=true,bookmarks=false]{hyperref}

\iccvfinalcopy 

\def\iccvPaperID{12386} 

\ificcvfinal\fi

\newcommand{\myrightarrow}{\ensuremath{\rightarrow}}

\begin{document}

\title{Handwritten and Printed Text Segmentation: A Signature Case Study\vspace{-2mm}}

\author{Sina Gholamian\\
\href{https://www.thomsonreuters.com/en/artificial-intelligence/research.html}{Thomson Reuters AI Labs}\\
Toronto, Canada\\
{\tt\small \href{mailto:sina.gholamian@thomsonreuters.com}{sina.gholamian@thomsonreuters.com}}
\and
Ali Vahdat\\
\href{https://www.thomsonreuters.com/en/artificial-intelligence/research.html}{Thomson Reuters AI Labs}\\
Toronto, Canada\\
{\tt\small \href{mailto:ali.vahdat@thomsonreuters.com}{ali.vahdat@thomsonreuters.com}}
}

\maketitle
\ificcvfinal\thispagestyle{empty}\fi

\begin{abstract}
While analyzing scanned documents, handwritten text can overlap with printed text. 
This overlap causes difficulties during the optical character recognition (OCR) and digitization process of documents, and subsequently, hurts downstream NLP tasks. 
Prior research either focuses solely on the binary classification of handwritten text or performs a three-class segmentation of the document, i.e., recognition of handwritten, printed, and background pixels. 
This approach results in the assignment of overlapping handwritten and printed pixels to only one of the classes, and thus, they are not accounted for in the other class. 
Thus, in this research, we develop novel approaches to address the challenges of handwritten and printed text segmentation. 
Our objective is to recover text from different classes in their entirety, especially enhancing the segmentation performance on overlapping sections.
To support this task, we introduce a new dataset, SignaTR6K, collected from real legal documents, as well as a new model architecture for the handwritten and printed text segmentation task. 
Our best configuration outperforms prior work on two different datasets by 17.9\% and 7.3\% on IoU scores. 
The SignaTR6K dataset is accessible for download via the following link: \url{https://forms.office.com/r/2a5RDg7cAY}. 
\end{abstract}

\vspace{-3mm}
\section{Introduction}
\begin{figure}
    \centering
        \fbox{\includegraphics[width=.95\linewidth]{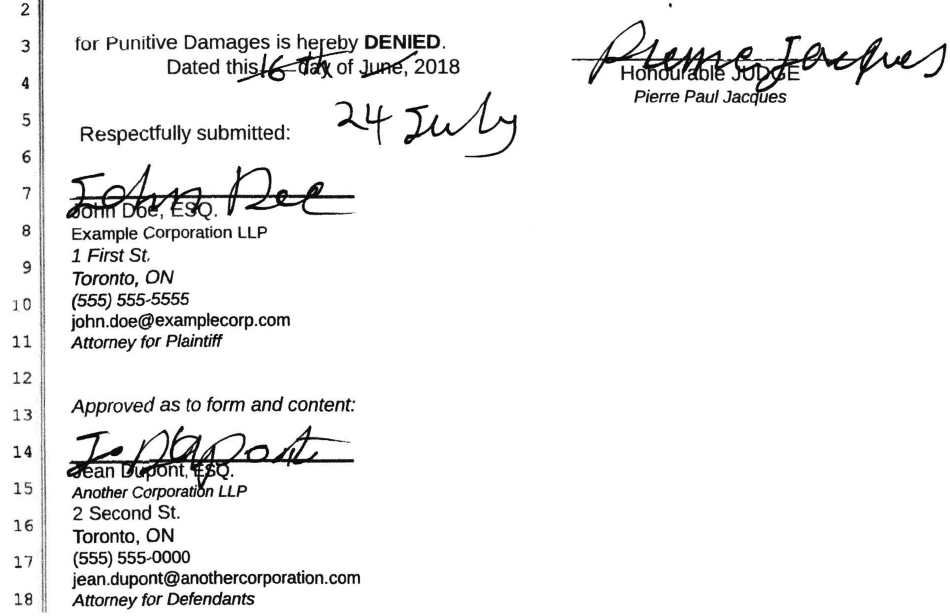}} 
    \caption{\small Court documents are first printed and then signed or annotated by various parties. This results in handwritten text overlapping the underlying printed information, leading to performance degradation in downstream tasks, such as named entity recognition (NER). This is a fabricated example, however very similar to the original documents, to protect personally identifiable information (PII).\looseness=-2}
    \label{fig:signature}
\vspace{-1mm}
\end{figure}

\vspace{-2mm}
For various purposes, the digitization of hard-copy documents and associated challenges are an active area of research in both academia~\cite{subramani2020survey,junior2020fcn+,vafaie2022handwritten,jo2020handwritten,dutly2019phti,prikhodina2021handwritten} and industry~\cite{norkute2021towards}. 
This digital transformation involves scanning paper documents through an OCR process, making their text accessible for downstream natural language processing (NLP) tasks, such as named entity recognition (NER). 
The documents of interest can originate from a variety of domains, including historical documents~\cite{vafaie2022handwritten}, legal and court-issued documents~\cite{norkute2021towards}, business contracts~\cite{subramani2020survey}, and medical records and prescriptions~\cite{dhar2021hp_docpres}. 
Although various studies, as cited above, have been conducted, there is yet a considerable gap between current approaches and human-level performance in mixed-text scenarios (i.e., when handwritten and printed text overlap). 
For instance, attorneys frequently sign legal documents, resulting in their signatures overlapping with their information. 
This overlap hampers the performance of OCR tools in character recognition, subsequently making it challenging for downstream tasks to accurately identify information linked to the attorneys and their associated law firms.
Figure~\ref{fig:signature} provides an illustration of handwritten text (HT) overlapping with printed text (PT) in court documents. 
Extracting parties names is a crucial step in the named-entity recognition (NER) task for legal and court documents~\cite{skylaki2020named}. 
When attorneys and other involved parties sign these documents, which are later scanned using OCR tools, their signatures often obscure the details of names and law firms. 
Consequently, the semantic segmentation of handwritten elements, such as lawyers' signatures and accompanying handwritten notes, and printed text detailing the lawyer and the law firm's information, becomes vital.\looseness=-1

In this research, we aim to address the challenges of HT and PT segmentation, as there is still a large gap between human-level performance and the existing approaches for this task. Our focus in this effort is to improve the segmentation performance in overlapping regions for scanned legal documents, and to aid in this endeavor, we also introduce a new dataset. 
In summary, our research makes the following contributions:\looseness=-1 
\begin{itemize}[leftmargin=*]
\item We introduce a new dataset, SignaTR6K (pronounce as Signature 6K)\footnote{Available at \url{https://forms.office.com/r/2a5RDg7cAY} .} derived from 200 pixel-level manually annotated crops of images from genuine legal documents. 
The dataset comprises signatures, handwritten text, and printed text, which frequently overlap. 
With data augmentation, we have created a dataset of sizes 5169, 530, 558, for training, validation, and testing, respectively, that we release to the public, to facilitate dataset availability for future research and to aid in the training and evaluation of deep-learning segmentation models.\looseness=-1\vspace{-1mm}

\item We propose a novel architecture that integrates both \textit{semantic segmentation features} and \textit{fine features}, enhancing the performance of text segmentation over previous methods. 
Moreover, we introduce a new loss function termed \textit{Fusion loss}, that, comparatively, is stable and converges to optimal loss values and Intersection over Union (IoU) scores.\looseness=0\vspace{-2mm} 

\item Lastly, we conduct an extensive quantitative and visual evaluation of different variations of our approach against prior work on two distinct datasets, and illustrate our approach's superior performance in the text segmentation task, especially in challenging scenarios where printed and handwritten text overlap. \vspace{-1mm} 
\end{itemize}

\section{Background and Literature Review}\label{rwork}
\vspace{-1mm}
The text segmentation problem is defined as follows: given a scanned document possibly containing handwritten, printed, and background (i.e., blank) pixels, the task is to assign each pixel to its appropriate class. 
Scanned documents can originate from various sources, such as hard-copy paper documents or microfilms~\cite{prikhodina2021handwritten,vafaie2022handwritten}.
Formally, for a given document $D$, assuming there exist three classes as handwritten text $HT$, printed text $PT$, and background $BG$, and pixel $p_i$:
\texttt{\hlblue{$\forall p_i \in D: p_i(c) == True \textit{\hspace{1mm}} if \textit{\hspace{1mm}}p_i \in c \textit{\hspace{1mm}}$ $\&\& \textit{\hspace{2mm}} c \in \{HT, PT, BG\}$}} \colorbox{cyan!10}{\hypertarget{form1}{(1)}}. 
A situation may arise where a pixel belongs to two classes, i.e., $HT$ and $PT$ classes, when handwritten text overlaps with printed text, which single-label three-class formulation cannot handle such cases.

Several previous studies have studied the segmentation of handwritten and printed text~\cite{dutly2019phti,jo2020handwritten,prikhodina2021handwritten,vafaie2022handwritten}, however, they have inherent limitations. 
Some focus solely on binary classification, determining if a pixel is handwritten or not~\cite{jo2020handwritten}, whereas others adopt a 3-class formulation of the problem, classifying pixels as handwritten, printed, or background~\cite{dutly2019phti,prikhodina2021handwritten,vafaie2022handwritten}. 
This exclusive assignment of pixels to three different classes paralyzes any machine learning segmentation model to properly detect pixels in the overlapping areas to belong to both handwritten and printed classes. 
In our usecase depicted in Figure~\ref{fig:signature}, the OCR process achieves insufficient performance due to overlapping of handwritten text and signatures that overlay the printed text. 
As such, to improve the documents digitization quality, and subsequently a wide range of downstream document understanding and natural language processing (NLP) tasks, it is vital to devise image processing approaches that can understand and properly segment different layers of text, i.e., printed and handwritten text.\looseness=-1 \vspace{-3mm}

\paragraph{Models.}
Prior literature has applied various approaches for the identification and separation of handwritten and printed text. 
Early approaches~\cite{franke1993writing,kandan2007robust} formulated the problem as a binary classification task using KNN and SVM and focusing on connected components (CCs) (i.e., groups of pixels). 
More recently, Li et al.~\cite{li2018printed} employed conditional random fields (CRFs), with formulating both unary and pairwise potentials for adjacent connected components by leveraging convolutional neural networks (CNNs) architecture for the separation of CCs. 
The limitation of CC-based approaches is that they determine the class membership for the entire component rather than at the pixel level. 
Consequently, pixel-level segmentation methods were introduced, leveraging Markov random fields (MRFs) and MLPs~\cite{peng2013handwritten,seuret2014pixel} for pixel-level classification of PT and HT segmentation. 
Following the success of encoder-decoder architectures in object segmentation~\cite{ronneberger2015u}, more recent works~\cite{jo2020handwritten,dutly2019phti,prikhodina2021handwritten,vafaie2022handwritten} have predominantly adopted a U-Net based architecture~\cite{ronneberger2015u}, which comprises an encoder-decoder network, for HT and PT segmentation. 

In addition to our contribution of releasing a manually annotated dataset of legal documents with overlapping text, SignaTR6K, our methodology distinguishes itself from prior works in several significant ways. 
Firstly, we approach the segmentation problem with a four-class formulation, allowing overlapping pixels to be assigned to a new distinct class (OV), which signifies the presence of both HT and PT layers, leading to enhanced segmentation performance. 
Secondly, we introduce a novel architecture, the Mixed Feature Model (MFM), that combines a Fine Feature Path (FFP) with a Semantic Segmentation Path (SSP) and improves the performance by capitalizing on both high-level and low-level features. 
Notably, existing U-Net style architectures~\cite{jo2020handwritten,dutly2019phti,prikhodina2021handwritten,vafaie2022handwritten} are limited to leveraging only the SSP path. 
Further, we present a new loss function, termed Fusion Loss, that converges faster and is more stable compared to prior losses for HT and PT segmentation task. 
In addition, we introduce a post-processing heuristic based on Conditional Random Fields (CRFH) to carry out relabeling, resulting in further enhancement of text segmentation performance. 
Additional details related to background and relevant works can be found in the supplementary.\looseness=-1
\vspace{-1mm}

\section{SignaTR6K Dataset}\label{sig6k}
\vspace{-1mm} 
Successfully training and testing a segmentation model requires access to high-quality labeled datasets. 
Due to the limited availability of public data that contains both handwritten and printed text, previous research has predominantly sought to synthetically generate such data. 
To achieve this, researchers have combined datasets that either exclusively contain printed or handwritten text, or those with non-overlapping text, such as IAM~\cite{marti2002iam}, RIMES~\cite{rimes}, PRImA~\cite{antonacopoulos2009realistic}, CVL~\cite{kleber2013cvl}, Scanned Questionnaire~\cite{qsnaire}, and WGM-SYN~\cite{vafaie2022handwritten}. 
The scanned documents can originate from diverse domains, each possessing its unique fonts, characters, and quality. 
Whether the original documents having poor quality in the case of archival documents~\cite{vafaie2022handwritten}, or the scanning process resulted in lower resolution or loss of contrast, these factors also compound the complexities of the text segmentation task. 
In addition, in some cases errors are introduced during the automated text labeling process~\cite{prikhodina2021handwritten}. 
Given these challenges, the absence of a precisely and manually annotated and verified dataset from real sources that contains various patterns of overlapping printed and handwritten text, significantly impedes the supervised machine learning (ML) process. 
To address this gap, we introduce a new dataset, \textbf{SignaTR6K} (pronounced as Signature 6K), that is derived from 200 original legal documents from Thomson Reuters Legal Content Services~\cite{trsite}. 
The dataset features overlapping printed and handwritten text and hand-drawn signatures, as depicted in Figure~\ref{fig:signature}. 
The documents originate from different organizations, including law firms and courts, each having its distinct fonts and document formats. 
Furthermore, the annotations are made by different individuals, ensuring a diverse range of printed and handwritten styles. 
Importantly, each document has been manually labeled and verified.
\begin{figure}\vspace{-1mm}
\captionsetup[subfigure]{aboveskip=0pt,belowskip=-1pt}
\centering
\begin{subfigure}{.15\textwidth}
  \centering
  \fbox{\includegraphics[width=\linewidth]{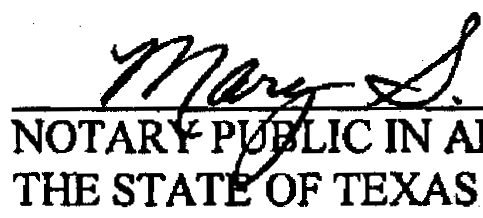}}
  \caption{\small Original crop}
  \label{fig:SigTR_example:sub1}
\end{subfigure}
\begin{subfigure}{.15\textwidth}
  \centering
  \fbox{\includegraphics[width=\linewidth]{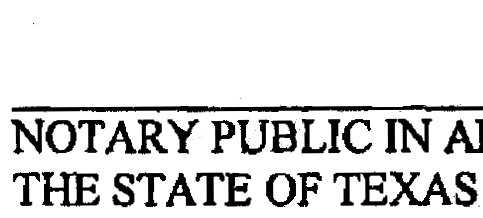}}
  \caption{\small Printed}
  \label{fig:SigTR_example:sub2}
\end{subfigure}
\begin{subfigure}{.15\textwidth}
  \centering
  \fbox{\includegraphics[width=\linewidth]{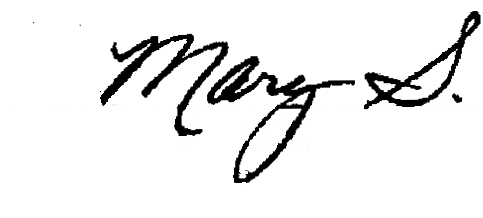}}
  \caption{\small Handwritten}
  \label{fig:SigTR_example:sub3}
\end{subfigure}
 \caption{\small This figure depicts a real crop from a legal document. (a) The original image containing printed text and overlaying handwritten signature; (b) Printed layer pixels annotated only; (c) Handwritten layer pixels annotated only. Any pixels not present in the printed or handwritten layers are marked as background.}
\label{fig:SigTR_example}
\end{figure}

Figure~\ref{fig:SigTR_example} displays an original crop from SignaTR6K, which includes both a signature and the information of the signing party. 
Due to the presence of personally identifiable information (PIIs), we cropped the images to ensure no visible PIIs. 
Figure~\ref{fig:SigTR_example:sub1} illustrates the original image with both printed text and overlaying handwritten signature. 
Figure~\ref{fig:SigTR_example:sub2} presents only the printed layer pixels, while Figure~\ref{fig:SigTR_example:sub3} contains only the handwritten layer pixels that are manually annotated. 
Any pixels absent from either the printed or handwritten channels are designated as background pixels.
It is also evident that in overlapping scenarios, some pixels belong to both handwritten and printed layers. \vspace{-3mm}

\begin{figure}
\captionsetup[subfigure]{aboveskip=-1pt,belowskip=-1pt}
\centering
\begin{subfigure}{.115\textwidth}
  \centering
  \fbox{\includegraphics[width=\linewidth]{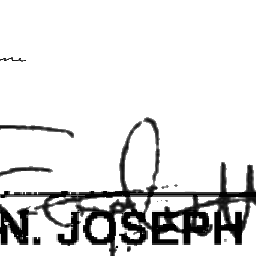}}
  \label{}
\end{subfigure}
\begin{subfigure}{.115\textwidth}
  \centering
  \fbox{\includegraphics[width=\linewidth]{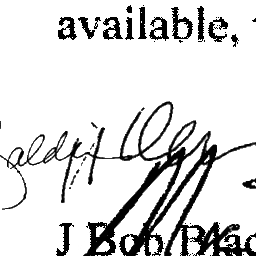}}
  \label{}
\end{subfigure}
\begin{subfigure}{.115\textwidth}
  \centering
  \fbox{\includegraphics[width=\linewidth]{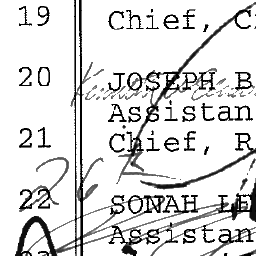}}
  \label{}
\end{subfigure}
\begin{subfigure}{.115\textwidth}
  \centering
  \fbox{\includegraphics[width=\linewidth]{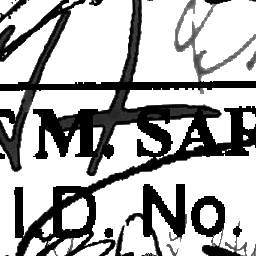}}
  \label{}
\end{subfigure}\vspace{-4mm}
\begin{subfigure}{.115\textwidth}
  \centering
  \fbox{\includegraphics[width=\linewidth]{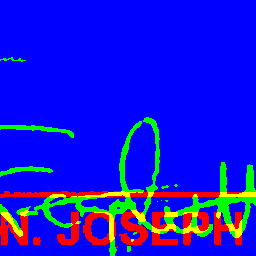}}
  \label{}
\end{subfigure}
\begin{subfigure}{.115\textwidth}
  \centering
  \fbox{\includegraphics[width=\linewidth]{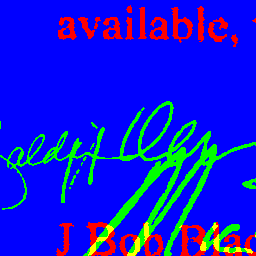}}
  \label{}
\end{subfigure}
\begin{subfigure}{.115\textwidth}
  \centering
  \fbox{\includegraphics[width=\linewidth]{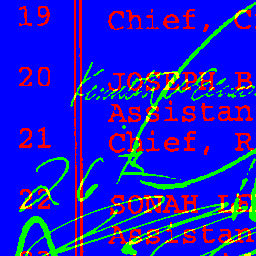}}
  \label{}
\end{subfigure}
\begin{subfigure}{.115\textwidth}
  \centering
  \fbox{\includegraphics[width=\linewidth]{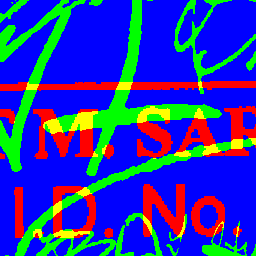}}
  \label{}
\end{subfigure}
\vspace{-8mm}
 \caption{Examples from the SignaTR6K dataset, with the top row showing the crops and the bottom row their ground truth annotations.  \textcolor{red}{Red}: class $PT$, \textcolor{red}{printed}, \textcolor{green}{Green}: class $HT$, \textcolor{green}{handwritten}, and \textcolor{blue}{Blue}: class $BG$, \textcolor{blue}{background}.}
\label{fig_sig_syn}
\end{figure}

\begin{figure*}[hbtp]
\vspace{-1mm}
    \centering
    \includegraphics[width=.9\linewidth]{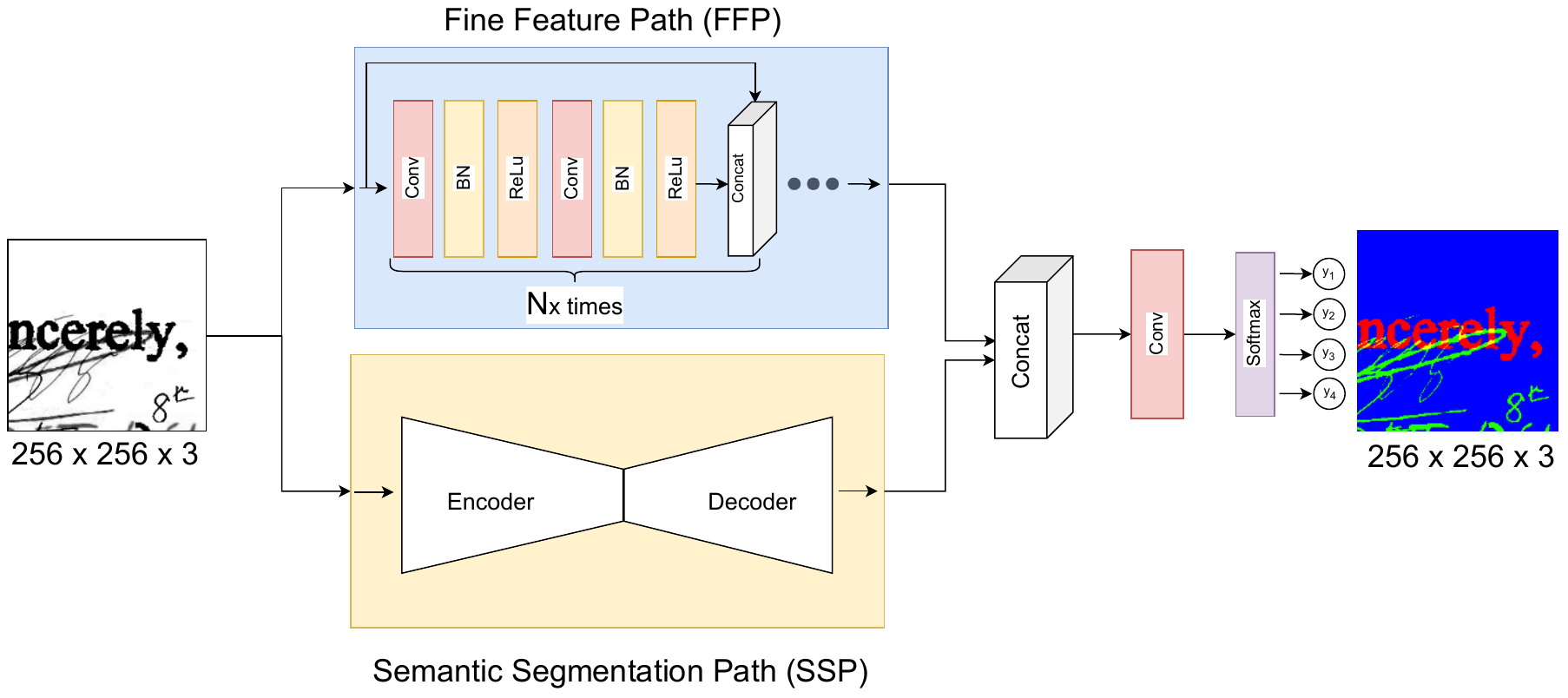} 
    \caption{Our architecture proposal showing FFP and SSP model paths. The output of FFP and SSP are concatenated prior to the final round of convolutions. A Softmax activation then selects one of the four classes for each pixel. The model's final output consists of pixel-level annotations of the input image. BN: BatchNorm.}
\label{fig:arch_full}
\end{figure*}

\vspace{-1mm}
\paragraph{Dataset Generation.} 
As hand-annotating a vast collection of real documents is time-consuming and expensive, in order to expand on the size of the dataset and make it adequate for training a deep learning model, we turn to data synthesis and augmentation techniques~\cite{jo2020handwritten}. 
The synthetic approaches include general augmentation of crops, along with shifting, magnifying, and rotating operations. 
We also overlay handwritten and printed pixels from different crops to generate new real-like examples of overlapping text. 
For this purpose, from the 200 distinct original document samples, we set aside 16 crops, ensuring mutually exclusive samples in the test set. 
These crops are augmented only for the test set,  and we create the training and validation sets from the remaining 184 samples. 
Following this approach and after excluding generated samples with visible PIIs, we have curated a dataset for training, validation, and testing sets of sizes 5169, 530, and 558 samples, respectively. Each sample is a pair of grayscale crop and its manually annotated ground truth. 
Each image is 256 by 256 pixels in size, with three channels (RGB) and typically contains several HT and PT overlaps. 
Four examples from the SignaTR6K dataset are presented in Figure~\ref{fig_sig_syn}, with their grayscale crops and corresponding ground truth (GT) labels. 
HT is indicated in green, PT in red, and BG in blue. 
Overlapping HT and PT pixels combine green and red channel values, resulting in a yellow appearance.
We envision this dataset can be utilized for model training from scratch or further fine-tuning of a pre-trained classification or segmentation model for specific tasks. 
The SignaTR6K dataset is freely available for public download.

\vspace{-1mm}
\section{Approach and Methodology}\label{approach}
\vspace{-1mm}
In this section, we detail our approach, the rationale behind the chosen architecture, the Semantic Segmentation Path (SSP) and Fine Feature Path (FFP), the various loss functions employed, and our novel \textit{Fusion} loss.
\vspace{-1mm}

\subsection{Model Architecture}
\vspace{-1mm}
The prevalent architecture for object segmentation applies a U-Net design~\cite{ronneberger2015u,long2015fully}. 
This architecture leverages a Fully Convolutional Network (FCN), that is without fully connected layers that are typically present at the end of Convolutional Neural Networks (CNNs). 
The U-Net architecture consists of two main parts, an encoder and a decoder. 
In the encoder segment, the original image (in our context a document crop) is fed into the model. It then undergoes a series of convolutions and max-pooling layers, extracting features from the image and reducing its dimensions. 
Conversely, the decoder processes the down-sampled image from the encoder using convolutions and up-sampling layers that eventually restores the image back to its original input size with the same number of channels. 
The final output of the decoder is a pixel-level labeling of the original image. 
Semantic Segmentation Path (SSP) in Figure~\ref{fig:arch_full} shows this part of the fully convolutional network. 
In addition to the encoder-decoder architecture in U-Net, there are also skip-connections that bring the feature maps directly from an encoder stage to the corresponding decoder stage, i.e., the decoder stage that has the same image size of the encoder stage. 
These skip-connections improve segmentation performance by re-accessing early-stage features that may be lost in the encoder's final output due to down-sampling.\looseness=-1

\vspace{-1mm}
\subsection{Four-Class Formulation}
\vspace{-1mm}
As mentioned earlier, prior approaches have either considered binary classification of HT or a three-class formulation of the problem. 
Binary classification detects only one type of text, while the three-class formulation results in overlapping pixels being assigned solely to either the HT or PT class, impairing performance. 
Therefore, we propose a four-class formulation of the segmentation task, and the fourth class, \textit{overlap}, $OV$, models pixels that belong to both the HT and PT classes, specifically enabling the detection of overlapping areas. 
To expand on Formula~(\hyperlink{form1}{1}) for this four-class formulation, for a given document $D$, assuming there exist four classes as handwritten text $HT$, printed text $PT$, background $BG$, and overlap $OV$, and pixel $p_i$: 
\hlblue{$\forall p_i \in D: p_i(c) == True$ $\textit{\hspace{1mm}} if \textit{\hspace{1mm}}p_i \in c \textit{\hspace{1mm}}$ $and$ $\textit{\hspace{2mm}} c \in \{HT, PT, BG, OV\}$}.


Overlapping pixels are highlighted in yellow in the ground truth, as seen in Figure~\ref{fig_sig_syn}. 
Our four-class single-label classification employs a Softmax activation function in the final layer, which ensures that only one output for each pixel is activated (Figure~\ref{fig:arch_full}). 
Since the output image comprises three channels, when the $OV$ class is predicted, during a post-processing step we turn on pixels for both $HT$ and $PT$ channels, resulting in the yellow color in the output image. 
In addition to the four-class formulation, we also explored three-class and multi-label formulations. 
In this scenario, instead of a Softmax activation, we applied Sigmoid activations for each class (i.e., three separate sigmoids). However, formulating the problem as a multi-label classification introduces added complexity and degrees of freedom. 
This can lead to undesirable scenarios, such as the simultaneous activation of pixels for $HT$ and $BG$. Consequently, the multi-label approach did not yield results comparable to those of the four-class formulation. 
  
\vspace{-1mm}
\subsection{Semantic Segmentation Path (SSP)}
\vspace{-1mm}
The semantic segmentation path (SSP) of our model leverages a U-Net based architecture, with down-sampling in the encoder stages (i.e., backbone) and up-sampling in the decoder stages. 
The U-net architecture works well in capturing high-level image features. 
In this architecture, the encoder and decoder maintain a symmetrical architecture with a similar number of stages. 
For example, in the SSP, if the encoder goes through four down-sampling stages, i.e., the input image size changes from 256*256 to 16*16 (256\myrightarrow 128\myrightarrow 64\myrightarrow 32\myrightarrow 16), correspondingly, the decoder undergoes four up-sampling stages, from 16*16 to 256*256. Thus, the final output retains the original input image's pixel dimensions. 
For the SSP, we explored a variety of network sizes. 
As a baseline and for comparison, we implemented FCN-light~\cite{dutly2019phti,prikhodina2021handwritten,vafaie2022handwritten}. 
We then improved on the SSP architecture by using VGG16~\cite{simonyan2014very}, InceptionV3~\cite{szegedy2016rethinking}, and ResNet34~\cite{he2016deep} as backbones of the SSP and observed improvement in the performance. 
In particular, the ResNet34 and InceptionV3 backbones outperform the prior work (FCN) due to their larger number of learnable parameters. 
Additionally, the inclusion of residual connections and varied convolution sizes allows to better carry over the low-level features of text to the later stages of the segmentation network. 
This observation further inspired us to introduce the FFP network, which similarly incorporates residual connections and without down-sampling.
 
\vspace{-1mm}
\subsection{Fine Feature Path (FFP)}
\vspace{-1mm}
The semantic segmentation path excels when segmenting distinct objects. 
In this path, down-sampling layers are rapidly applied to the input image to capture high-level features and patterns. 
However, this rapid processing can lead to the loss of fine, or low-level, features. 
For our application, low-level features are crucial due to the intertwined nature of printed and handwritten text. 
To address this, we introduce a parallel path to the SSP, termed the fine feature path (FFP), which avoids down-sampling and instead incorporates a convolution block with residual connections. 
Note that, while the FFP aids in capturing fine features without down-sampling, on its own, i.e., without the SSP that includes down-sampling, it is insufficient for text segmentation, as the absence of high-level features means the model will not be able to detect high-level patterns irrespective of their pixel locations in the image. 
In Figure~\ref{fig:arch_full}, the fine feature block of the FFP is repeated $N_x$ times; in our implementation, $N_x=4$. 
In addition, each stage of the FFP itself implements residual connections as it provides the flexibility to either bypass the block or use its output, leading to improved results, as we will discuss later in the paper. 
Similar architectures employing residual blocks have shown improved performance in fine object segmentation tasks, e.g., road segmentation~\cite{zhang2018road}.     
Table~\ref{table:finefeature} details the FFP architecture for four stages ($N_x=4$) and the connections between layers and convolution sizes.

\begin{table}
	\vspace{-1mm}
	\tiny
	\centering
	\resizebox{.5\textwidth}{!}{
	\begin{tabular}{cclllll}
		\toprule
		& Group  & Layer type & Filter  &  Input(s) & Output(s) &Output size\\
		\hline	
		\multirow{5}{*}{Stage 1}
		& \multirow{3}{*}{G 1}& Conv  & $3\times 3/64$ & img\_input  & s1\_g1\_c\_o &  $256\times 256 \times 64$ \\
		&        & BatchNorm  &  \_ &  s1\_g1\_c\_o & s1\_g1\_b\_o & $256\times 256 \times 64$ \\
		&        & ReLu  &  \_ & s1\_g1\_b\_o & s1\_g1\_r\_o &  $256\times 256 \times 64$ \\
		\cline{2-7}			
		& \multirow{4}{*}{G 2}& Conv  &  $3\times 3/64$ & s1\_g1\_r\_o & s1\_g2\_c\_o&  $256\times 256 \times 64$ \\
		&& BatchNorm  & \_ & s1\_g2\_c\_o & s1\_g2\_b\_o &  $256\times 256 \times 64$ \\
		&  & Relu  &  \_ &  s1\_g2\_b\_o & s1\_g2\_r\_o & $256\times 256 \times 64$ \\
		&  & Concat  & \_ & img\_in, s1\_g2\_r\_o & s1\_o &  $256\times 256 \times 67$ \\		
		\hline

		\multirow{7}{*}{Stage 2}
		& \multirow{3}{*}{G 1}
		&Conv  & $3\times 3/64$ &  s1\_o & s2\_g1\_c\_o &  $256\times 256 \times 64$ \\
		&& BatchNorm  & \_ & s2\_g1\_c\_o & s2\_g1\_b\_o &  $256\times 256 \times 64$ \\
		&  & Relu  &  \_ &  s2\_g1\_b\_o & s2\_g1\_r\_o & $256\times 256 \times 64$ \\
			\cline{2-7}
		&\multirow{4}{*}{G 2}		
	    &Conv  &  $3\times 3/64$ & s2\_g1\_r\_o & s2\_g2\_c\_o&  $256\times 256 \times 64$ \\
    	&& BatchNorm  &  \_ &  s2\_g2\_c\_o & s2\_g2\_b\_o & $256\times 256 \times 64$ \\
		&  & ReLu  &  \_ & s2\_g2\_b\_o & s2\_g2\_r\_o &  $256\times 256 \times 64$ \\
    	& & Concat  & \_ & s1\_o, s2\_g2\_r\_o & s2\_o &  $256\times 256 \times 131$ \\		
		\hline	
		
		\multirow{7}{*}{Stage 3}
		& \multirow{3}{*}{G 1}
		&Conv  & $3\times 3/64$ & s2\_o & s3\_g1\_c\_o &  $256\times 256 \times 64$ \\
		&& BatchNorm  & \_ & s3\_g1\_c\_o & s3\_g1\_b\_o &  $256\times 256 \times 64$ \\
		&  & Relu  &  \_ &  s3\_g1\_b\_o & s3\_g1\_r\_o & $256\times 256 \times 64$ \\
		\cline{2-7}
		&\multirow{4}{*}{G 2}		
		& Conv  &  $3\times 3/64$ & s3\_g1\_r\_o & s3\_g2\_c\_o&  $256\times 256 \times 64$ \\

		&& BatchNorm  &  \_ &  s3\_g2\_c\_o & s3\_g2\_b\_o & $256\times 256 \times 64$ \\
		&  & ReLu  &  \_ & s3\_g2\_b\_o & s3\_g2\_r\_o &  $256\times 256 \times 64$ \\
	
		&        & Concat  & \_ & s2\_o, s3\_g2\_r\_o & s3\_o &  $256\times 256 \times 195$ \\		
		\hline

		\multirow{7}{*}{Stage 4}
		& \multirow{3}{*}{G 1}
		&Conv  & $3\times 3/64$ &  s3\_o & s4\_g1\_c\_o &  $256\times 256 \times 64$ \\
		&& BatchNorm  & \_ & s4\_g1\_c\_o  & s4\_g1\_b\_o &  $256\times 256 \times 64$ \\
		&  & Relu  &  \_ &  s4\_g1\_b\_o & s4\_g1\_r\_o & $256\times 256 \times 64$ \\
				\cline{2-7}
		&\multirow{4}{*}{G 2}
		&Conv  &  $3\times 3/64$ & s4\_g1\_r\_o & s4\_g2\_c\_o&  $256\times 256 \times 64$ \\
		&& BatchNorm  &  \_ &  s4\_g2\_c\_o & s4\_g2\_b\_o & $256\times 256 \times 64$ \\
		&  & ReLu  &  \_ & s4\_g2\_b\_o & s4\_g2\_r\_o &  $256\times 256 \times 64$ \\
	
	   &  & Concat  & \_ & s3\_o, s4\_g2\_r\_o & s4\_o &  $256\times 256 \times 259$ \\		
		\hline	

		Output& \_     & Conv & $1\times 1/4$ & s4\_o  & FFP\_o &   $256\times 256 \times 4$ \\		\bottomrule
	\end{tabular}}
	\caption{The architecture of the Fine Feature Path (FFP). In our design, we use four stages (Stages 1-4) of fine feature blocks. We include BatchNorm and ReLu activation after each convolution layer, and unlike SSP, down-sampling layers (e.g., max pooling) are absent from FFP. The output of each stage is concatenated with its input to create residual connections.}
	\label{table:finefeature}	
\end{table}
\subsection{Mixed Feature Model (MFM)}
\vspace{-1mm}
The Mixed Feature Model (MFM) serves as the umbrella model containing two parallel paths: SSP and FFP, as depicted in Figure~\ref{fig:arch_full}.   
The objective of MFM is to capture the image's low-level features alongside its high-level features. 
The outputs of SSP and FFP are concatenated before the final layer convolution, producing the output of the MFM model. 
Additional details regarding the architecture, inputs, outputs, and convolution layer sizes of the MFM can be found in the supplementary.

\subsection{CRF Post-Processing and CRF Heuristic}\label{crfh}
\vspace{-1mm}
Prior research~\cite{dutly2019phti,prikhodina2021handwritten,vafaie2022handwritten} has leveraged dense Conditional Random Fields (CRFs) as a post-processing step to re-label pixels based on their neighboring pixels. In our architectural exploration, we found that, while CRF post-processing is intended to improve segmentation performance, in practice it often hurts the segmentation performance by aggressively re-labeling pixels to incorrect classes. 
One issue arises due to the imbalanced nature of pixels across classes, with background pixels being predominant. 
Consequently, many pixels are mistakenly annotated from HT or PT to BG, or from HT to PT, which is undesirable. 
Based on this observation, we designed a heuristic for CRF post-processing to contain this unfavourable behaviour by only allowing the BG pixels to be relabelled to HT or PT pixels and not vice versa. 
As we will discuss in the experimentation section, this heuristic further improves segmentation performance.\looseness=-2

\subsection{Loss Functions}
Due to the nature of the scanned documents and the amount of text on each page, the number of background pixels (i.e., white blank pixels) surpasses that of handwritten and printed pixels. 
This leads to a class imbalance problem~\cite{zhou2005training}, posing the risk of predicting the majority of pixels as background while still minimizing the loss value. 
As a result, we investigated various loss functions and weight assignments to different classes in the loss function to evaluate its impact on segmentation performance. 
Additionally, during this exploration, we observed that different loss functions achieve different IoU (Intersection over Union) scores. 
Consequently, we introduce a new loss function, Fusion Loss, to incorporate the benefits of various losses. 
In the following, we discuss the different loss functions we explored.\looseness=-1

\paragraph{Cross-Entropy Loss.} The multi-class cross-entropy loss is used for classification tasks involving more than two classes. It assumes that for each data point, only one class can be the ground truth, i.e., multi-class single-label classification. 
The standard form of cross-entropy loss for a single data point in multi-class segmentation with $M$ classes is defined as: \hlblue{$\mathcal{L}_{CE}(gt, pr) = - \sum_{m=1}^{M} gt \cdot \log(pr)$}, where $pr$ and $gt$ represent the prediction and ground truth, respectively. 
The final loss is the average of the losses for all data points in a training set. 
Similarly, the weighted version (WCE) is computed as \hlblue{$\mathcal{L}_{WCE}(gt, pr) = - \sum_{m=1}^{M} w_m \cdot gt \cdot \log(pr)$}, with $w_m$ being the weight assigned to each class in the loss calculation.



\paragraph{Focal Loss.} The focal loss~\cite{lin2017focal} was introduced to address the class-imbalance problem during training of object detection tasks. 
The focal loss puts the focus of the learning algorithm on the incorrectly classified examples by applying a modulating term \hlblue{$(1-pr)^{\gamma}$} to the cross-entropy loss. 
This scaling factor dynamically weighs down the contribution of easy and correctly classified samples, allowing the training loop to concentrate on the harder and incorrectly classified samples. 
The focal loss is calculated as: \hlblue{$\mathcal{L}_{Focal}(gt, pr) = - \sum_{m=1}^{M} gt \cdot (1-pr)^{\gamma} \log(pr)$}, where $\gamma$ is the modulating/focusing factor, and we set \hlblue{$\gamma == 2$} as per the original paper. 
Similarly, the weighted version of the focal loss can be expressed as: \hlblue{$\mathcal{L}_{WF}(gt, pr) = - \sum_{m=1}^{M} gt \cdot \alpha_m \cdot (1-pr)^{\gamma} \log(pr)$}, where $\alpha_m$ is analogous to $w_m$, assigning weights to each class such that: \hlblue{$\forall  \alpha_m\, \mid \, 0 < \alpha_m <1\ \&\& \  \sum_{m=1}^{M} \alpha_m = 1$}.

%

\vspace*{-2mm}
\paragraph{Dice Loss.}
Dice loss is given by \hlblue{$\mathcal{L}_{Dice}(precision, recall)$ $= 1$ $- \frac{2}{M}\sum_{m=1}^{M}$ $\frac{ precision_m \cdot recall_m}{precision_m + recall_m}$}. 
Intuitively, it aims to maximize the F-Score while minimizing the loss, i.e., \hlblue{$\mathcal{L}_{Dice} = 1 - F_{Score}$}. 
Accordingly, the weighted version of the dice loss is defined as \hlblue{$\mathcal{L}_{WD}(precision, recall)$ $  = 1- \frac{2}{M}$ $\sum_{m=1}^{M}  \frac{w_m \cdot precision_m \cdot recall_m}{precision_m + recall_m}$}.



\paragraph{Fusion Loss.}
Given our observation that different loss functions adversely affect different classes, we introduce a new loss targeted for maximizing the performance of all the classes. 
For example, the dice loss performs better on the background class, as in its formulation, it aims to maximize the F-Score. 
Because the majority of pixels are attributed to the background class, having higher values of correctly classified background pixels achieves a higher F-Score and lower loss values. 
However, this does not necessarily yield higher IoU scores for PT and HT classes. 
In contrast, the weighted versions of the cross-entropy and focal losses focus on the handwritten and printed classes. 
As such, with the Fusion loss, we aim to combine the behaviors of various losses, and we define the Fusion loss as the sum of the three weighted losses:  
\colorbox{myblue}{$\mathcal{L}_{Fusion}= \mathcal{L}_{WF} + \mathcal{L}_{WCE} +\mathcal{L}_{WD}$}.    



\begin{table*}
\resizebox{\linewidth}{!}{%
\begin{tabular}{@{}cccllccc|lccc|lccc}
\toprule

&&  && \multicolumn{4}{c}{\textbf{IoU \%}}       &  \multicolumn{4}{c}{\textbf{With CRF (IoU \%)}} &  \multicolumn{3}{c}{\textbf{With CRFH (IoU \%)}}     \\  \cmidrule(lr){5-8} \cmidrule(l){9-12} \cmidrule(l){13-16} 
Formulation                                                                          & Backbone & \# Parameters & Loss function  & PT & HT & BG& Mean &  PT & HT & BG & Mean  &  PT & HT & BG & Mean \\ \midrule
\multirow{2}{*}{\textbf{\begin{tabular}[c]{@{}c@{}}3-Class\end{tabular}}}  
& \textbf{FCN-light}~\cite{dutly2019phti}& $\sim$295K& Weighted CE &  24.00& 26.00& 72.00& 41.00& 11.00& 23.00&72.00&36.00&\_&\_&\_&\_ \\
&\textbf{WGM-MOD}~\cite{vafaie2022handwritten}&$\sim$295K& Weighted CE&42.00&36.00&74.00&50.00\textsuperscript{*}&41.00&32.00&74.00&49.00&\_&\_&\_&\_ \\
\midrule 
\multirow{28}{*}{\textbf{\begin{tabular}[c]{@{}c@{}} \Large 4-Class\\(Ours)\end{tabular}}}  
& \multirow{7}{*}{\textbf{\begin{tabular}[c]{@{}c@{}}FCN-light\end{tabular}}}  
&\multirow{7}{*}{{\begin{tabular}[c]{@{}c@{}}$\sim$295K\end{tabular}}} 

&CE&46.49&41.98&73.77&54.08&38.95&29.35&73.92&47.41&46.57&41.59&73.87&54.01\\
&&&Focal&46.02&42.01&73.80&53.95&32.96&24.50&74.00&43.82&46.11&41.23&73.98&53.77\\
&&& Dice&48.01&47.28&71.22&55.55&48.59&48.48&71.22&56.10&48.07&47.79&71.29&55.72\\
&&& Weighted CE&46.07&42.18&73.82&54.02&40.54&30.07&73.98&48.20&46.31&41.68&73.95&53.98\\
&&& Weighted Focal&43.71&41.77&73.95&53.14&32.67&24.40&74.07&43.71&44.06&41.12&74.11&53.10\\
&&& Weighted Dice&47.97&47.22&71.02&55.40&48.69&48.34&71.02&56.02&48.04&47.70&71.10&55.61\\
&&& Fusion&48.12&47.25&71.93&55.77&46.34&44.25&71.86&54.15&48.19&47.89&72.38&56.15\\

\cmidrule(l){2-16} 
&\multirow{7}{*}{\textbf{\begin{tabular}[c]{@{}c@{}}SSP - VGG16\end{tabular}}}
&\multirow{7}{*}{{\begin{tabular}[c]{@{}c@{}}$\sim$24M\end{tabular}}} 
&CE&35.14&32.40&73.96&47.17&30.97&19.14&73.88&41.33&35.21&32.08&74.13&47.14\\
&&&Focal&34.56&32.22&74.02&46.93&29.11&19.42&73.91&40.81&34.67&31.79&74.22&46.89\\
&&&Dice&40.64&39.64&71.13&50.47&43.30&40.16&71.21&51.55&40.81&40.25&71.25&50.77\\
&&& Weighted CE&35.73&33.44&74.43&47.87&32.00&20.70&74.12&42.28&35.93&32.95&74.65&47.84\\
&&& Weighted Focal&34.90&34.35&74.36&47.87&29.18&20.72&74.06&41.32&35.11&33.80&74.56&47.82\\
&&& Weighted Dice&41.80&40.64&70.67&51.03&44.14&41.39&70.70&52.07&42.52&41.19&70.97&51.56\\
&&& Fusion&39.99&39.56&72.31&50.62&35.39&27.93&72.35&45.22&40.10&39.94&72.63&50.89\\

\cmidrule(l){2-16} 
&\multirow{7}{*}{\textbf{\begin{tabular}[c]{@{}c@{}}MFM (FFP + SSP) - InceptionV3\end{tabular}}}
&\multirow{7}{*}{{\begin{tabular}[c]{@{}c@{}}$\sim$30M\end{tabular}}} 
&CE&52.51&44.09&73.91&56.84&46.03&35.94&74.21&52.06&52.55&42.17&74.35&56.36\\
&&&Focal&51.56&43.74&73.89&56.40&35.89&27.15&74.41&45.82&51.63&41.74&74.43&55.93\\
&&& Dice&20.91&19.28&\underline{81.52}&40.57&25.51&24.61&\underline{79.94}&43.35&20.92&19.29&\textbf{\underline{81.54}}&40.58\\
&&& Weighted CE&51.58&43.72&73.95&56.42&44.78&34.04&74.32&51.05&51.64&41.85&74.40&55.96\\
&&& Weighted Focal&51.01&43.62&73.89&56.18&35.76&26.60&74.22&45.53&51.08&41.60&74.35&55.68\\
&&& Weighted Dice&22.30&21.25&{80.73}&41.42&29.11&29.48&77.97&45.52&22.32&21.26&{{80.74}}&41.44\\
&&& Fusion&\underline{{53.22}}&51.25&71.84&58.77&50.05&48.83&72.03&56.97&\underline{\textbf{53.32}}&49.83&72.72&58.62\\

\cmidrule(l){2-16} 
&\multirow{7}{*}{\textbf{\begin{tabular}[c]{@{}c@{}} MFM (FFP + SSP) - ReNet34\end{tabular}}}
&\multirow{7}{*}{{\begin{tabular}[c]{@{}c@{}}$\sim$24M\end{tabular}}} 
&CE&52.40&44.02&73.91&56.78&47.52&36.40&74.12&52.68&52.43&42.12&74.34&56.30\\
&&&Focal&51.95&43.98&73.89&56.61&36.73&27.17&74.31&46.07&51.90&41.99&74.33&56.08\\
&&& Dice&29.68&29.43&77.06&45.39&31.62&25.39&75.15&44.05&29.84&29.63&77.07&45.52\\
&&& Weighted CE&52.63&44.12&73.94&56.90&47.60&37.25&74.12&52.99&52.56&42.22&74.36&56.38\\
&&& Weighted Focal&51.68&43.95&73.90&56.51&35.73&27.22&74.30&45.75&51.72&41.98&74.34&56.01\\
&&& Weighted Dice&51.44&50.60&71.96&58.00&\underline{52.97}&\underline{51.42}&71.90&\underline{58.76}&51.54&49.37&72.75&57.89\\
&&& Fusion&{52.99}&\underline{\textbf{51.72}}& 71.69&\underline{58.80}&51.15&50.79&71.83&57.92&53.01&\underline{51.29}&72.49&\underline{\textbf{58.93}}\\
\bottomrule
\end{tabular}%
}
    \captionof{table}{IoU performance (\%) on the WGM-SYN dataset~\cite{vafaie2022handwritten}. The maximum value of a column (representing a class) is underlined. The overall maximum of a class across different post-processing methods is  both bolded and underlined. For example, the best mean IoU for the WGM-SYN dataset is achieved with Fusion loss, using CRFH, and the MFM-ResNet34 architecture, stands at \underline{\textbf{58.93}}. The best performing configuration from prior work, denoted with (\textsuperscript{*}), stands at 50.00.}
    \label{tab:res_wgm}
\end{table*}

\section{Experiments and Results}\label{exp_setup}
\vspace{-1mm}
\paragraph{Evaluation Metric.}
Intersection over Union (IoU) is a  commonly used metric to measure the performance of a segmentation task~\cite{wang2020image}, and in particular text segmentation~\cite{subramani2020survey,jo2020handwritten,dutly2019phti,vafaie2022handwritten}. 
In our context, we use IoU to measure the pixel-level performance of the model output versus the ground truth. 
For a class $c$, let $TP_c$ represent the correctly predicted pixels for that class, $FP_c$ denote the pixels incorrectly predicted as belonging to class $c$, and $FN_c$ represent the pixels that are incorrectly not predicted for class $c$, then Intersection over Union, IoU, for class $c$ is calculated as: \colorbox{myblue}{$IoU_c = \frac{TP_c}{TP_c+FP_c+FN_c}$}\vspace{1mm}.
For each experiment, we calculate the IoU for all three classes, HT, PT, and BG. 
In the four-class formulation, pixels attributed to the overlap (OV) class are converted in a post-processing step to both HT and PT classes, contributing to the IoU calculations for both. 
We also calculate the mean IoU, which is the average of IoUs for these three classes: \colorbox{myblue}{\vspace{1.5mm}$IoU_{Mean} = \frac{IoU_{HT}+IoU_{PT}+IoU_{BG}}{3}$}.\vspace{-3mm}

\paragraph{Experiments.}
We perform 50 epochs of training with a batch size of 8. 
For problem formulation, we use both three-class and four-class implementations. 
The experiments are conducted on two datasets: SignaTR6K and WGM-SYN~\cite{vafaie2022handwritten}. 
The IoU values for FCN-light and WGM-MOD are as reported in~\cite{vafaie2022handwritten} for the three-class formulation which has been performed on the same dataset, WGM-SYN. 
However, for the SignaTR6K dataset, we retrain an FCN-based model for three-class formulation to ensure a fair comparison on the new dataset.
We run experiments for a variety of architectures and in three post-processing configurations. 
We structure experiments incrementally, i.e., adding one improvement at a time. 
This approach aids in understanding and isolating the impact of each added improvement, i.e., ablation study. 
Due to space constraints, the table detailing the complete list of experiment parameters and configurations is relegated to the supplementary. 
  
\vspace{-2mm}
\paragraph{Experimented Architectures.}
We start with FCN-light which has approximately 295K parameters. 
This architecture aligns with prior research and serves as our baseline~\cite{dutly2019phti,vafaie2022handwritten}. 
Next, we improve the performance of FCN-light by adopting a four-class implementation. 
We then experiment with SSP architecture with three different backbones: VGG16, InceptionV3, and ResNet34. 
Our experimentation also includes three variations of MFM, which incorporates both FFP and SSP, and with three SSP variations: VGG16, InceptionV3, and ResNet34.        
For our implementations, we use the Segmentation Models Library available in TensorFlow~\cite{Yakubovskiy:2019}. \vspace*{-3mm}

\paragraph{Post-Processing Configurations.}
We experiment with three distinct configurations: without post-processing, with CRF post-processing, and with CRFH post-processing. 
In the scenario without post-processing, the IoU calculation is performed directly on the output of the model (FCN-light, SSP, and MFM). 
For the CRF post-processing approach, all classes are permissible for relabeling. 
However, with the CRF post-processing with heuristic (CRFH), as explained in Section~\ref{crfh}, only the background class pixels (BG) are allowed to be relabeled as HT or PT classes. 


\paragraph{Weight Initialization.}
For both FCN-light and SSP models, model weights start from their initial random values at the beginning of the training. 
We train MFM configurations at last to reuse the weights from the SSP configurations. 
Thus, for the MFM trainings, we apply transfer learning by initializing with weights from the corresponding, previously trained SSP. 
For example, when training an MFM (FFP+SSP) with an SSP using the ResNet34 backbone, we initialize its SSP path with the available weights from the SSP - ResNet34. 
Lastly, although we have aimed for consistent IoU results by fixing the random seeds, the inherent non-determinism associated with GPU execution remains. 
As such, we have run our experiments multiple times to validate the IoU value improvements across configurations.
\vspace{-2mm}

\begin{table*}
\vspace{-1mm}
\resizebox{\linewidth}{!}{%
\begin{tabular}{@{}cccllccc|lccc|lccc}
\toprule
&&  && \multicolumn{4}{c}{\textbf{IoU \%}}       &  \multicolumn{4}{c}{\textbf{With CRF (IoU \%)}} &  \multicolumn{3}{c}{\textbf{With CRFH (IoU \%)}}     \\  \cmidrule(lr){5-8} \cmidrule(l){9-12} \cmidrule(l){13-16} 
Formulation                                                                          & Backbone & \# Parameters & Loss function  & PT & HT & BG& Mean &  PT & HT & BG & Mean  &  PT & HT & BG & Mean \\ \midrule
\multirow{7}{*}{\textbf{\begin{tabular}[c]{@{}c@{}}3-Class\end{tabular}}}  
& \multirow{7}{*}{\textbf{\begin{tabular}[c]{@{}c@{}}FCN-based~\cite{dutly2019phti,vafaie2022handwritten}\end{tabular}}}  & \multirow{6}{*}{{\begin{tabular}[c]{@{}c@{}}$\sim$295K\end{tabular}}} 
& CE &62.56&88.09&98.40&83.02\textsuperscript{*}&52.68&89.68&99.26&80.46&62.72&90.58&99.05&84.11 \\
&&& Focal &62.34&88.00&98.45&82.93&44.60&84.86&99.28&76.25&62.57&90.83&99.21& 84.21\\
&&&Dice&60.45&87.29&97.85&81.86&61.24&88.83&98.17&82.74&60.52&87.88&97.97&82.12\\
&&&Weighted CE&60.58&84.89&97.78&81.09&53.45&90.73&99.52&81.23&60.84&86.33&98.24&81.80\\
&&&Weighted Focal&61.25&85.74&98.02&81.67&44.27&85.77&99.50&76.51&61.55&87.42&98.58&82.52\\
&&&Weighted Dice &60.21&87.00&97.74&81.65&60.72&88.17&97.98&82.29&60.27&87.46&97.83&81.86\\
&&&Fusion &61.52&88.39&98.38&82.76&58.94&92.55&99.37&83.62&61.67&90.45&98.91&83.68\\
\midrule 
\multirow{21}{*}{\textbf{\begin{tabular}[c]{@{}c@{}}\Large 4-Class\\(Ours)\end{tabular}}}  
& \multirow{7}{*}{\textbf{\begin{tabular}[c]{@{}c@{}}FCN-based\end{tabular}}}  
&\multirow{7}{*}{{\begin{tabular}[c]{@{}c@{}}$\sim$295K\end{tabular}}} 
&CE &64.55&89.21&98.39&84.05&54.60&89.68&99.23&81.17&64.87&91.81&99.06&85.25\\
&&&Focal &64.10&88.86&98.32&83.76&46.64&86.01&99.26&77.30&64.34&91.65&99.11&85.03\\
&&&Dice &64.37&88.68&98.37&83.81&65.17&90.13&98.59&84.63&64.57&89.14&98.46&84.06\\
&&&Weighted CE &63.78&87.52&98.16&83.15&54.77&90.50&99.42&81.56&64.12&89.43&98.71&84.09\\
&&&Weighted Focal&63.68&87.71&98.19&83.20&48.18&87.10&99.45&78.24&64.05&89.72&98.80&84.19\\
&&&Weighted Dice &64.21&88.57&98.30&83.69&65.08&90.10&98.53&84.57&64.44&89.06&98.39&83.96\\
&&&Fusion &64.68&88.48&98.33&83.83&59.83&91.77&99.28&83.63&65.00&90.72&98.91&84.87\\
\cmidrule(l){2-16} 

      &\multirow{7}{*}{\textbf{\begin{tabular}[c]{@{}c@{}} MFM (FFP + SSP) - InceptionV3\end{tabular}}}
      &\multirow{7}{*}{{\begin{tabular}[c]{@{}c@{}}$\sim$30M\end{tabular}}} 
      &CE &73.10&\underline{92.66}&98.77&\underline{88.18}&63.48&92.89&99.55&85.31&73.05&94.89&99.36&\underline{\textbf{89.10}}\\
      &&&Focal &72.80&92.50&98.75&88.01&57.47&90.58&99.54&82.53&72.77&94.74&99.35&88.95\\
      &&&Dice &72.56&92.20&98.70&87.82&72.38&93.60&99.00&88.33&72.52&94.73&99.37&88.87\\
      &&&Weighted CE &72.66&91.90&98.54&87.70&62.65&93.04&\underline{\textbf{99.62}}&85.10&72.63&93.31&98.94&88.29\\
      &&&Weighted Focal &72.55&92.13&98.62&87.77&59.35&91.64&99.59&83.53&72.50&93.77&99.07&88.45\\
      &&&Weighted Dice &72.59&92.31&98.70&87.87&72.60&94.00&99.07&\underline{88.56}&72.54&94.60&99.32&88.82\\
      &&&Fusion &72.55&92.25&98.71&87.83&65.57&93.49&99.53&86.19&72.49&94.62&99.35&88.82\\

\cmidrule(l){2-16} 
&\multirow{7}{*}{\textbf{\begin{tabular}[c]{@{}c@{}}MFM (FFP + SSP) - ResNet34\end{tabular}}}
&\multirow{7}{*}{{\begin{tabular}[c]{@{}c@{}}$\sim$24M\end{tabular}}} 
&CE &72.81&92.56&\underline{98.78}&88.05&63.04&92.94&99.55&85.17&72.75&\underline{\textbf{94.93}}&\underline{{99.39}}&89.02\\
&&&Focal &73.04&92.46&98.75&88.08&53.02&89.04&99.55&80.54&73.00&94.75&99.35&89.03\\
&&&Dice &72.96&92.38&98.72&88.02&\underline{73.16}&93.35&98.93&88.48&72.93&94.79&99.36&89.03\\
&&&Weighted CE &72.96&91.96&98.69&87.87&64.49&92.47&99.56&85.51&72.90&94.19&99.27&88.79\\
&&&Weighted Focal &73.18&92.10&98.63&87.97&54.96&89.71&99.60&81.42&73.16&93.72&99.06&88.65\\
&&&Weighted Dice &72.78&92.32&98.71&87.94&72.95&93.66&99.00&88.53&72.75&94.67&99.34&88.92\\
&&&Fusion &\underline{\textbf{73.26}}&92.45&98.73&{88.15}&68.38&\underline{94.85}&99.56&87.60&\underline{73.21}&94.59&99.31&{89.04}\\
                                                                     
\bottomrule
\end{tabular}%
}
    \captionof{table}{IoU performance (\%) on the SignaTR6K dataset. The maximum value of a column (i.e., class) is underlined, and the overall maximum for a class across different post-processing is both bolded and underlined. For example, the best mean IoU for the SignaTR6K dataset is for CE loss, with CRFH, and the MFM-InceptionV3 architecture at \underline{\textbf{89.10}}. The best performing configuration of prior work, i.e., excluding results for Fusion loss and CRFH, is marked with (\textsuperscript{*}) and is 83.02.}

    \label{tab:signaTR6K}
   \vspace{3mm}
\end{table*}

\paragraph{Results.}\label{results}
Tables~\ref{tab:res_wgm} and~\ref{tab:signaTR6K} show the IoU values for the WGM-SYN and SignaTR6K datasets. 
The overall trend across the results indicates that transitioning from the three-class formulation to the four-class formulation improves the IoU scores. 
Furthermore, employing larger model backbones generally, but not always, improves the segmentation performance. 
Among all the model architectures evaluated, the ResNet34 and InceptionV3 backbones achieve the highest performance, which we attribute to their residual connections and different size convolutions as they can better capture fine features from the image. 
We provide a summary of the trends in the results here, while a more detailed version of the result tables can be found in the supplementary material.

\begin{itemize}
 \setlength\itemsep{-1mm}
\item Prior work~\cite{dutly2019phti,vafaie2022handwritten}, which implements the three-class formulation of HT and PT segmentation, generally exhibits the lowest performance. 
Transitioning from the three-class to the four-class formulation with the same backbone, i.e., FCN-light, with the same number of model parameters and WCE loss, improves the average IoU values by 8.0\%  (50\myrightarrow 54.02) for the WGM-SYN dataset,  and 2.6\% (81.09\myrightarrow 83.15) for the SignaTR6K dataset.\looseness=-2 
\item Applying CRF post-processing generally degrades the results. 
In our research, we observe that CRF post processing is dataset dependent and decreases the performance of segmentation in some cases. CRF post-processing relabels aggressively and incorrectly converts HT and PT pixels to background ones, or OV pixels to PT ones. For example, Figure~\ref{fig:i} illustrates that how OV pixels are wrongly relabeled to PT class by CRF post-processing. 
Contrary to CRF post-processing, our CRF heuristic generally improves the IoU values. 
For example, for the SignaTR6K dataset, for FCN-light with four-class implementation and CE loss, CRFH improves HT IoUs by 2.9\% (89.21\myrightarrow 91.81).   

\begin{figure}
\vspace{-3mm}
\captionsetup[subfigure]{aboveskip=-1pt,belowskip=-1pt}
\centering
\begin{subfigure}{.23\textwidth}
  \centering
\includegraphics[width=1.1\linewidth]{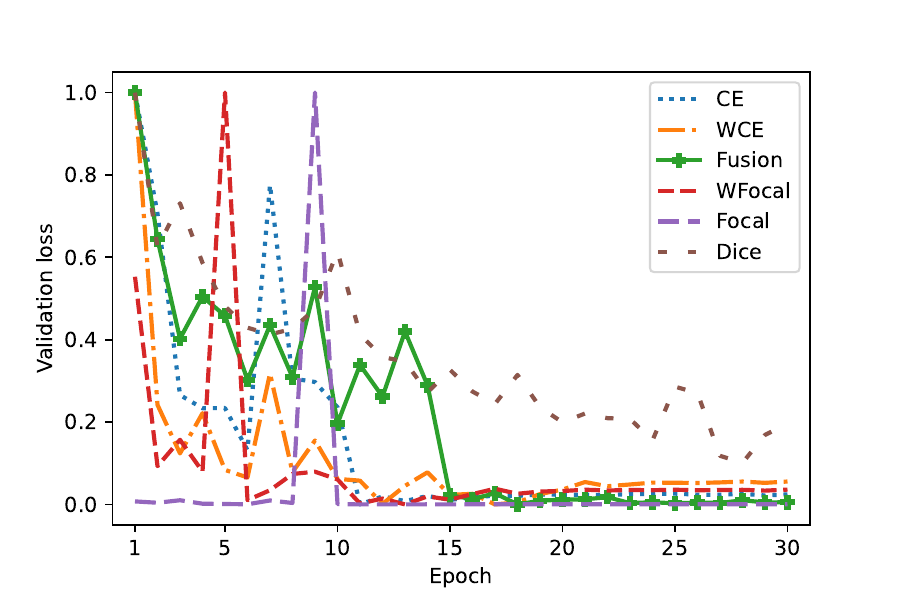} 
  \caption{Validation set losses}
  \label{fig:loss_vals}
\end{subfigure}
\begin{subfigure}{.23\textwidth}
  \centering
\includegraphics[width=1.1\linewidth]{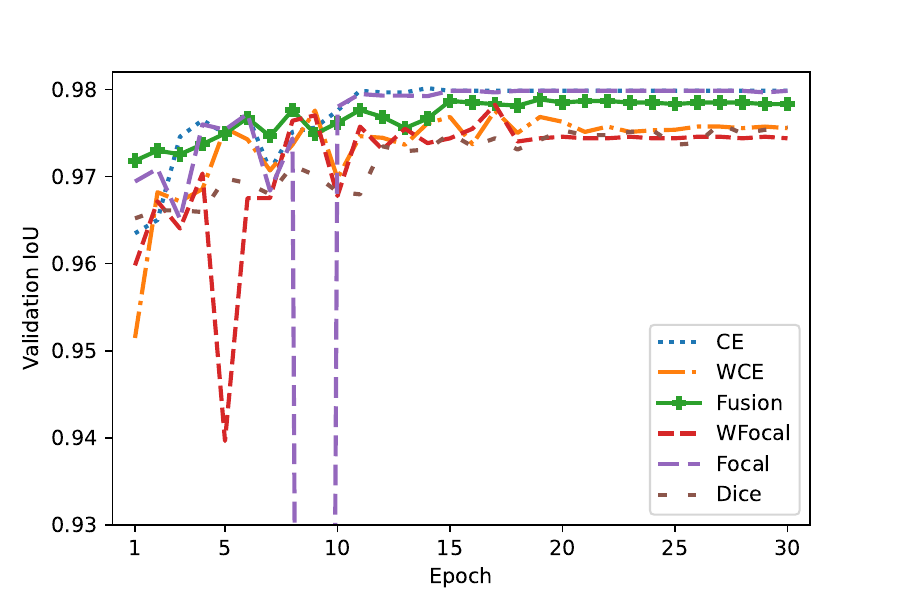} 
  \caption{Validation set IoUs}
  \label{fig:conv_speed}
\end{subfigure}
\caption{Convergence speed of different loss functions and normalized loss values on the validation set of the SignaTR6K dataset.}
\label{fig:losses}
\end{figure}

\item Moving from the FCN-light architecture to larger models, i.e., SSP and MFM models, generally improves the results, except for the VGG16 backbone. 
For VGG16, we rationalize that having a deep network without residual connections or varied-size convolutions, as seen in ResNet34 and InceptionV3 architectures, hurts the model's ability in capturing fine features. 
In addition, we observe that adding FFP helps to improve the performance of the VGG model as we compare SSP and MFM performance values for both datasets. 
This also confirms the FFP's ability to capture low-level features that are missed by the SSP path.

\begin{figure*}
\captionsetup[subfigure]{aboveskip=-1pt,belowskip=-1pt}
\centering
\begin{subfigure}{.12\textwidth}
  \centering
  \fbox{\includegraphics[width=\linewidth]{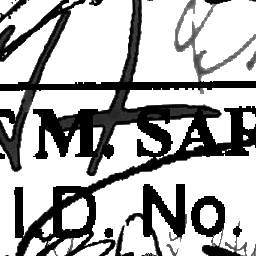}}
  \caption{}
  \label{tr200:a}
\end{subfigure}
\begin{subfigure}{.12\textwidth}
  \centering
  \fbox{\includegraphics[width=\linewidth]{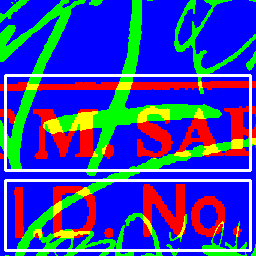}} 
  \caption{}
  \label{tr200:b}
\end{subfigure}
\begin{subfigure}{.12\textwidth}
  \centering
  \fbox{\includegraphics[width=\linewidth]{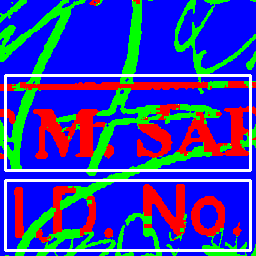}}
  \caption{}
  \label{tr200:c}
\end{subfigure}
\begin{subfigure}{.12\textwidth}
  \centering
  \fbox{\includegraphics[width=\linewidth]{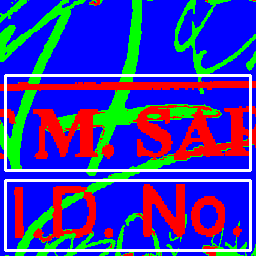}} 
  \caption{}
  \label{tr200:d}
\end{subfigure}
\begin{subfigure}{.12\textwidth}
  \centering
  \fbox{\includegraphics[width=\linewidth]{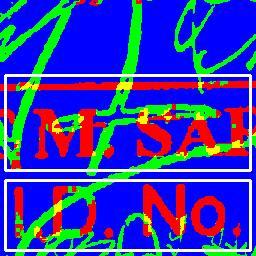}}
  \caption{}
  \label{tr200:e}
\end{subfigure}
\begin{subfigure}{.12\textwidth}
  \centering
  \fbox{\includegraphics[width=\linewidth]{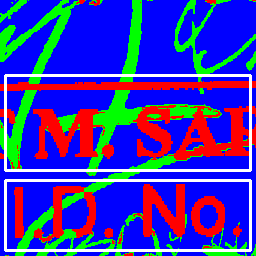}} 
  \caption{}
  \label{tr200:f}
\end{subfigure}
\begin{subfigure}{.12\textwidth}
  \centering
  \fbox{\includegraphics[width=\linewidth]{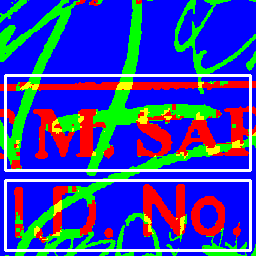}} 
  \caption{}
  \label{tr200:g}
\end{subfigure}
\begin{subfigure}{.12\textwidth}
  \centering
  \fbox{\includegraphics[width=\linewidth]{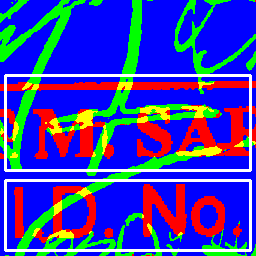}} 
  \caption{}
  \label{tr200:h}
\end{subfigure}

\begin{subfigure}{.12\textwidth}
  \centering
  \fbox{\includegraphics[width=\linewidth]{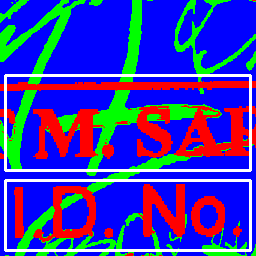}} 
  \caption{}
  \label{fig:i}
\end{subfigure}
\begin{subfigure}{.12\textwidth}
  \centering
  \fbox{\includegraphics[width=\linewidth]{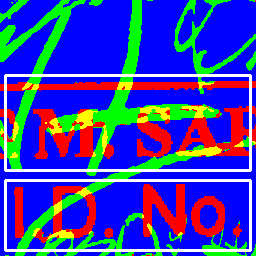}} 
  \caption{}
  \label{tr200:j}
\end{subfigure}
\begin{subfigure}{.12\textwidth}
  \centering
  \fbox{\includegraphics[width=\linewidth]{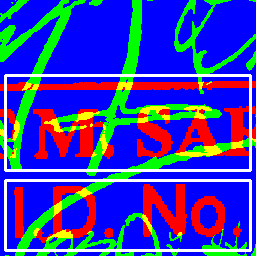}} 
  \caption{}
  \label{tr200:k}
\end{subfigure}
\begin{subfigure}{.12\textwidth}
  \centering
  \fbox{\includegraphics[width=\linewidth]{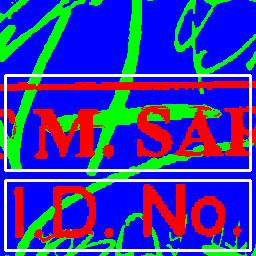}} 
  \caption{}
  \label{tr200:l}
\end{subfigure}
\begin{subfigure}{.12\textwidth}
  \centering
  \fbox{\includegraphics[width=\linewidth]{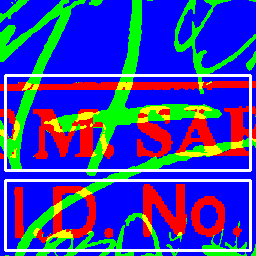}} 
  \caption{}
  \label{tr200:m}
\end{subfigure}
\begin{subfigure}{.12\textwidth}
  \centering
  \fbox{\includegraphics[width=\linewidth]{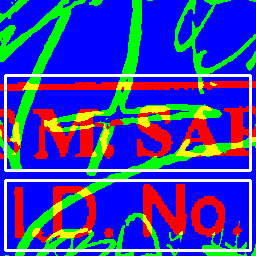}} 
  \caption{}
  \label{tr200:n}
  \end{subfigure}
\begin{subfigure}{.12\textwidth}
  \centering
  \fbox{\includegraphics[width=\linewidth]{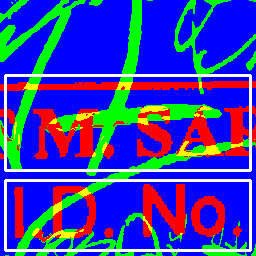}} 
  \caption{}
  \label{tr200:o}
\end{subfigure}
\begin{subfigure}{.12\textwidth}
  \centering
  \fbox{\includegraphics[width=\linewidth]{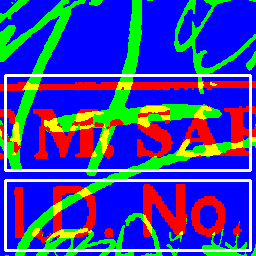}}
  \caption{}
  \label{tr200:p}
\end{subfigure}

\caption{Example results on the test set of the SignaTR6K dataset for our approach compared to the ground truth and prior works. (a) Input image; (b) Ground truth; (c) \& (d) 3-class FCN-based~\cite{dutly2019phti,vafaie2022handwritten} with CE loss without (c) and with (d) CRF post-processing; (e), (f), \& (g) Our FCN-based 4-class formulation with CE loss without CRF (e), with CRF (f), and with CRFH (g); (h), (i), \& (j) SSP-ResNet34 with CE loss without CRF (h), with CRF (i), and with CRFH (j); (k), (l), \& (m) MFM-ResNet34 with CE loss without CRF (k), with CRF (l), and with CRFH (m); (n), (o), \& (p) MFM-ResNet34 with Fusion loss without CRF (n), with CRF (o), and with CRFH (p).}
\label{fig:tr200}
\vspace{1mm}
\end{figure*}

\item The MFM model with ResNet34 and InceptionV3 backbones generally achieves the best results. 
MFM-ResNet34 with Fusion loss and CRFH achieves the best result for the WGM-SYN dataset (58.93\%), while MFM-InceptionV3 with CE loss and CRFH achieves the best mean IoUs for the SignaTR6K dataset (89.10\%). 
Overall, the best-performing model from our designs improves on the mean IoU performance of the best-performing prior work~\cite{vafaie2022handwritten} by 17.9\% (50\myrightarrow 58.93) and 7.3\% (83.02\myrightarrow 89.10) for the WGM-SYN and SignaTR6K datasets, respectively.   

\item Although using skip connections in SSP helps with the segmentation performance
and thus less pronounced improvement is observed in MFM, our FFP design is agnostic of the SSP implementation. 
This distinction becomes clear when VGG16 is employed as SSP; the addition of FFP boosts the IoU scores by 7\% and 8\% on the WGM-SYN and SignaTR6K datasets, respectively. 
The FCN-based model with 295K parameters outperforms VGG16 with 30M parameters. 
Thus, larger models do not necessarily outperform smaller ones consistently, and integrating FFP can bring significant improvements in IoU scores. Our findings show that, irrespective of the SSP size and architecture, incorporating FFP improves the performance, and it is essential for segmentation tasks involving fine objects, e.g., text.
\end{itemize}
\vspace{-2mm} 
\vspace{-2mm}
\paragraph{Loss Functions.}
We also perform an analysis to confirm the stability and convergence of the Fusion loss compared to other losses. 
Figures~\ref{fig:loss_vals} and~\ref{fig:conv_speed} compare different loss functions on the validation set of the SignaTR6K dataset. 
In Figure~\ref{fig:loss_vals}, we observe that the Fusion loss is generally stable while converging to its minimum loss.
Additionally, in Figure~\ref{fig:conv_speed}, we observe Fusion loss reaches its maximum IoU around Epoch 15, generally faster than most other losses and stays close to the maximum IoU values on the validation set. 
It is important to highlight that the maximum IoU on the validation set for Focal and CE losses does not necessarily imply the best performance on the test set. 
Overall, Fusion loss shows a stable behavior and achieves a better performance on the test set, whereas for comparison, Focal loss shows instability at some epochs during the training.
\vspace{-2mm}
\paragraph{Visual Comparison.}
Figure~\ref{fig:tr200} shows sample model outputs for the SignaTR6K dataset, with two rectangle regions of interest highlighted to showcase the differences between various models. 
Figure~\ref{tr200:b} shows the ground truth, and a visual trend indicates that the performance on the regions of interest improves from Figure~\ref{tr200:c} to Figure~\ref{tr200:p}. 
It is also visually noticeable that CRF post-processing aggressively relabels pixels (\ref{fig:i}), whereas CRFH generally improves the results. 
More visual comparisons on the WGM-SYN and SignaTR6K datasets can be found in the supplementary.

\vspace{-2mm}
\paragraph{Limitations.} 
Compared to prior work, the complexity and training cost of our approach present some limitations.
Our MFM model, being larger than FCN-light, requires greater GPU memory sizes for training, and takes longer to train. 
However, we believe some of the limitations on the training time can be offset through transferring weights from SSP models to MFM ones. 
Additionally, while our approach outperforms prior work on the WGM-SYN dataset, the mean IoU performance is lower compared to the SignaTR6K dataset. 
This indicates potential limitations in our method's efficacy for lower-quality original documents, those undergoing low-quality scanning processes (like historical documents), or those with errors from automated labeling. 
We also attribute the improved performance on the SignaTR6K dataset to its higher quality and our thorough manual annotation, that have resulted in better model training and improved IoU results.

\section{Conclusion}\label{conc}
\vspace{-1mm}
Segmentation of handwritten text (HT) and printed text (PT) is vital for digitization and understanding of scanned documents. 
The complexity increases with the overlapping of different text types. 
In this research, we introduced SignaTR6K, a new open-source dataset with high-quality, manual pixel-level annotations, sourced from original legal documents.
Additionally, we proposed a novel four-class formulation and a new architecture for the segmentation task. 
Our design leverages both the Fine Feature Path (FFP) and the Semantic Segmentation Path (SSP) to create the Mixed Feature Model (MFM), and incorporates both high-level and low-level features and improves on the text segmentation performance. 
We also introduced a CRF-based post-processing heuristic (CRFH) that further improves the model output, and included a new loss function, Fusion loss, that combines the advantages of different loss functions and achieves faster convergence and higher stability compared to most of the evaluated losses. 
In conclusion, our designs outperform the prior work in mean IoU scores by 17.9\% and 7.3\% on WGM-SYN and SignaTR6K datasets, respectively.

\clearpage
\bibliographystyle{ieee_fullname}
\bibliography{iccv_bib}

\appendix
\section*{{\Large Appendices}}
In this section,  we provide further background and additional details on the experiments, architecture, and results for our work on handwritten and printed text segmentation. 
\section{Additional Background and Related Work}
\subsection{Scope}
Converting hard-copy documents, including microfilms for archival and historical documents~\cite{vafaie2022handwritten}, into digital format has been the focus of the computer vision research community, and several approaches have been proposed for this purpose~\cite{islam2017survey,subramani2020survey}. 
For a variety of reasons, such as ease of access and document understanding, paper documents in different domains, such as books, archival documents, medical records, legal documents, survey forms, and more, are converted to their digital format through optical character recognition (OCR). 
Due to the high demand for document digitization, several commercial and open-source OCR engines, such as Amazon's Textract~\cite{amazon_textract}, Tesseract~\cite{tesseract}, and ABBYY FineReader~\cite{abby} have been available for some time. 
These tools significantly speed up the automatic digitization of documents at scale. 
However, given the diversity of document types, their structures, overlapping handwritten and printed text, and also the poor quality of original documents and their scanned version in the case of historical documents, the quality of OCR tools can degrade in the digitization process~\cite{vafaie2022handwritten,garlapati2017system}. 
Consequently, our approach seeks to tackle and improve the quality of text segmentation to benefit the downstream tasks.

\begin{table*}[!htb]
	\centering
	\begin{tabular}{cclllll}
		\toprule
		& Group  & Layer type & Filter  &  Input(s) & Output(s) &Output size\\
		\hline
Input   &\_     &Input&  \_ &   img\_input & i\_o   &  $256\times 256 \times 3$ \\
	  \hline 
		\multirow{3}{2cm}{Fine Feature Path}
		& \multirow{3}{*}{G 1}
		 & 	FFP  &  \_ &  i\_o & FFP\_o & $256\times 256 \times 4$\Tstrut \\
		&  & BatchNorm  &  \_ & FFP\_o & g1\_b\_o &  $256\times 256 \times 4$ \\
	   && ReLu  & \_ & g1\_b\_o & g1\_r\_o &  $256\times 256 \times 4$ \\
     \hline
     
     	\multirow{3}{2cm}{Semantic Segmentation Path}
		& \multirow{3}{*}{G 2}
		 & 	SSP  &  \_ &  i\_o & SSP\_o & $256\times 256 \times 4$\Tstrut \\
		&  & BatchNorm  &  \_ & SSP\_o & g2\_b\_o &  $256\times 256 \times 4$ \\
	   && ReLu  & \_ & g2\_b\_o & g2\_r\_o &  $256\times 256 \times 4$ \\
     \hline

     \multirow{3}{2cm}{Concatenation}
		& \multirow{3}{*}{G 3}
		 & 	Concat  &  \_ &   g1\_r\_o, g2\_r\_o & g3\_cc\_o & $256\times 256 \times 8$\Tstrut \\
	& & Conv& $1\times 1/4$ & g3\_cc\_o & g3\_cv\_o &  $256\times 256 \times 4$ \\
	   && Softmax  & \_ & g3\_cv\_o & g3\_s\_o &  $256\times 256 \times 4$ \\
     \hline
	Output& \_     & Output& \_ & g3\_s\_o  & $MFM_{output}$ &   $256\times 256 \times 4$ \\		

	\bottomrule
	\end{tabular}
    \vspace{1mm}
	\caption{\small The architecture and connections of the high-level model, Mixed Feature Model (MFM).}	
	\label{table:total_model}	
	\vspace{2mm}
\end{table*}

\subsection{Models}
Parts of this section were presented in the main text, and in the following, we offer a more comprehensive review. 
Prior work has employed various methods for handwritten and printed text segmentation. 
Early approaches~\cite{franke1993writing,kandan2007robust} treated the problem as a binary classification task using classifiers like KNN and SVM. 
They utilized connected components (CCs) (i.e., group of pixels), and various sets of features to determine whether a CC represents handwritten (HT) or printed (PT) text. 
These features include geometric features of CCs such as their heights, widths, and the spread between CCs~\cite{franke1993writing}, as well as geometric-invariant features such as invariant moments~\cite{kandan2007robust}. 
More recently, Li et al.~\cite{li2018printed} applied conditional random fields (CRFs), with formulating both unary and pair-wise potentials for adjacent connected components by leveraging convolutional neural networks (CNNs) architecture for the separation of CCs. 
The primary limitation of CC-based approaches is that they assign a single class for the entire component, rather than assigning classes at the pixel-level. 
Additionally, they fail to detect overlapping regions since the entire connected component is categorized as either HT or PT.

Due to the drawbacks of connected components, pixel-level segmentation methods were introduced that leverage Markov random fields (MRFs)~\cite{peng2013handwritten}. 
The authors of~\cite{peng2013handwritten} applied both patch-level and pixel-level classification for PT and HT segmentation. 
These classifications were initially identified using a G-means based approach (a modified version of k-means~\cite{hamerly2003learning}), followed by a relabeling step based on MRFs. 
Seuret et al.~\cite{seuret2014pixel} classified the foreground pixels as either printed or handwritten text using an MLP architecture with two fully-connected hidden units. 
This MLP was followed by a post-processing step designed to correct probable mistakes based on adjacent pixels. 

\begin{figure}[h]
\captionsetup[subfigure]{aboveskip=1pt,belowskip=1pt}
\centering
\begin{subfigure}{.23\textwidth}
 \centering
 \fbox{\includegraphics[width=\linewidth]{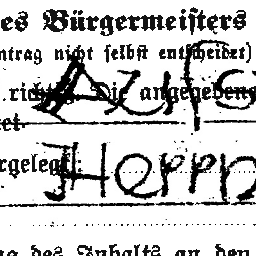}}
 \caption{Gray-scale}
 \label{}
\end{subfigure}
\begin{subfigure}{.23\textwidth}
 \centering
 \fbox{\includegraphics[width=\linewidth]{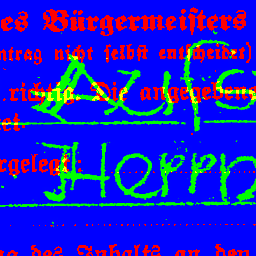}}
 \caption{Ground truth}
 \label{}
\end{subfigure}
\caption{An example from the WGM-SYN dataset~\cite{vafaie2022handwritten}. \textcolor{red}{Red}: class $PT$, \textcolor{red}{printed}, \textcolor{green}{Green}: class $HT$, \textcolor{green}{handwritten}, and \textcolor{blue}{Blue}: class $BG$, \textcolor{blue}{background}.
}
\label{fig:wgm}
\vspace{2mm}
\end{figure} 

\begin{figure*}[h]
	\vspace{2mm}
    \centering
    \includegraphics[width=1.0\linewidth]{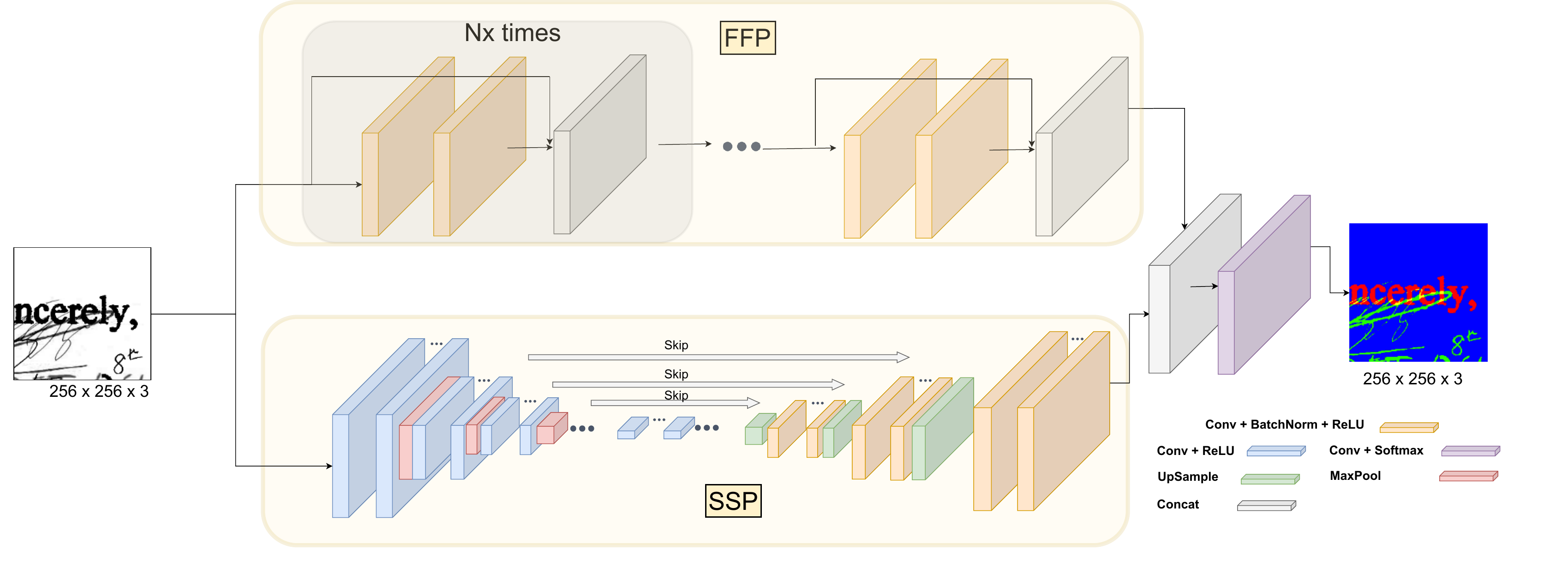} 
    \caption{Our proposed architecture uses the fine feature path (FFP) in parallel with the U-Net architecture (SSP) to capture low-level image features, while the U-Net captures high-level features through a condensing and expanding pipeline. In SSP, the three dots in the middle indicate that different sets of encoders and decoders can be applied. As such, we explored various backbones, including FCN-light, VGG16, InceptionV3, and ResNet34.}
    \label{fig:arch_detail} \vspace{1mm}
\end{figure*}

As encoder-decoder architectures have proven to perform well in object segmentation~\cite{ronneberger2015u}, recent works~\cite{vafaie2022handwritten,jo2020handwritten,dutly2019phti,prikhodina2021handwritten} have predominantly applied a U-Net based architecture~\cite{ronneberger2015u} for HT and PT segmentation. 
This architecture consists of an encoder-decoder design, similar to SSP shown in Figure~\ref{fig:arch_detail}. 
Jo et al~\cite{jo2020handwritten} utilized a U-Net architecture to perform binary classification of handwritten text. 
Similarly, authors in several studies~\cite{vafaie2022handwritten,dutly2019phti,prikhodina2021handwritten} leveraged a fully convolutional network (FCN) to classify three classes: handwritten (HT), printed (PT), and background (BG). 
The approaches are adapted for different applications such as web-based services~\cite{dutly2019phti} and understanding of historical documents~\cite{vafaie2022handwritten,prikhodina2021handwritten}. 
They also incorporate a conditional random field (CRF) post-processing step to re-label pixels based on their adjacent majority pixels. 
These approaches adhere to a three-class formulation of text segmentation, which assigns overlapping pixels to either the HT or PT classes. 
Moreover, the CRF post-processing often performs aggressive re-labeling that degrades the segmentation performance~\cite{vafaie2022handwritten,prikhodina2021handwritten}.

\subsubsection{WGM-SYN Dataset}
Besides the SignaTR6K dataset, we also performed evaluations on the WGM-SYN dataset~\cite{vafaie2022handwritten}. 
This dataset contains a subset of historical and archival records and documents from the ``Pilotprojekt zur Wiedergutmachung'' archive~\cite{wiedergutmachung}. 
The dataset is comprised of forms, typewritten certificates, declarations, and testimonies with different layouts, both in color and grayscale. 
Then, the documents were manually annotated for different text types with VOTT3~\cite{vott}. 
This process resulted in 319 images of handwritten and 767 images of machine-written text, both from microfilm and document scans. 
Finally, after some preprocessing, noise removal, and binarization steps, data synthesis techniques adopted from~\cite{jo2020handwritten} are applied. 
This process yields final training, validation, and testing set of sizes 3335, 430, and 430, respectively. 
Each data sample consists of a grayscale image crop of size 256*256 pixels containing both handwritten and printed text, along with a color-coded label: handwritten in green, printed in red, and background in  blue. 
Figure~\ref{fig:wgm} illustrates an example from the WGM-SYN dataset.

\begin{table}[h!]
    \small
    \centering
     \begin{tabular}{p{1.7cm}|p{6cm}} 
     \toprule
     Parameter & Details \\ [0.5ex] 
     \midrule\midrule
     \rowcolor{Gray}Training epochs & 50  \\ 
     Batch size & 8 \\ 
     \rowcolor{Gray}Loss functions& Cross-entropy, WCE, Focal, WF, Dice, WD, and Fusion\\ 
	 Weighted Loss functions weights& 3-class: [PT, HT, BG] = [0.4, 0.5, 0.1], 4-class: [PT, HT, BG, OV]=[0.3, 0.3, 0.1, 0.3]\\ 
     \rowcolor{Gray}Initial learning rate (LR) & 0.001\\     
     LR schedule & LR = LR/10 if no reduction in validation loss for 4 epochs.\\ 
     \rowcolor{Gray}Initial weights& FCN: None; SSP: None; MFM: SSP initialized with prior training weights\\ 
      Optimizer& Adam\\
      \rowcolor{Gray}SignaTR6K&Training: 5169; Validation: 530; Test: 558\\
      WGM-SYN&Training: 3335; Validation: 430; Test: 430\\
       \rowcolor{Gray}Problem formulation& three-class and four-class\\
       Architecture variations & WGM-MOD \& FCN-light (3-class); FCN-light (4-class); SSP (VGG16, InceptionV3, and Resnet34); MFM (VGG16, InceptionV3, and ResNet34)\\
     [1ex] 
     \bottomrule
     \end{tabular}
       \caption{\small Experimentation parameters.}	
      \label{table:parameters}	
    \end{table}

\section{Architecture} 
In this section, we provide additional details on the overall architecture of MFM in 
Table~\ref{table:total_model} outlines the stages and connections for MFM, and Figure~\ref{fig:arch_detail} offers a detailed depiction of the SSP and FFP blocks.

\section{Additional Experiments and Results}

\subsection{Experiment Configuration}
Table~\ref{table:parameters} provides additional details on the experimented configurations. For learning rate (LR), we start with the initial value of 0.001 and divide by ten if there is no improvement in validation loss values after four epochs.

\begin{table*}
\resizebox{\linewidth}{!}{%
\begin{tabular}{@{}cccllccc|lccc|lccc}
\toprule

&&  && \multicolumn{4}{c}{\textbf{IoU \%}}       &  \multicolumn{4}{c}{\textbf{With CRF (IoU \%)}} &  \multicolumn{3}{c}{\textbf{With CRFH (IoU \%)}}     \\  \cmidrule(lr){5-8} \cmidrule(l){9-12} \cmidrule(l){13-16} 
Formulation                                                                          & Backbone & \# Parameters & Loss function  & PT & HT & BG& Mean &  PT & HT & BG & Mean  &  PT & HT & BG & Mean \\ \midrule
\multirow{2}{*}{\textbf{\begin{tabular}[c]{@{}c@{}}3-Class\end{tabular}}}  
& \textbf{FCN-light}~\cite{dutly2019phti}& $\sim$295K& Weighted CE &  24.00& 26.00& 72.00& 41.00& 11.00& 23.00&72.00&36.00&\_&\_&\_&\_ \\
&\textbf{WGM-MOD}~\cite{vafaie2022handwritten}&$\sim$295K& Weighted CE&42.00&36.00&74.00&50.00\textsuperscript{*}&41.00&32.00&74.00&49.00&\_&\_&\_&\_ \\
\midrule 
\multirow{49}{*}{\textbf{\begin{tabular}[c]{@{}c@{}} \Large 4-Class\\(Ours)\end{tabular}}}  
& \multirow{7}{*}{\textbf{\begin{tabular}[c]{@{}c@{}}FCN-light\end{tabular}}}  
&\multirow{7}{*}{{\begin{tabular}[c]{@{}c@{}}$\sim$295K\end{tabular}}} 

&CE&46.49&41.98&73.77&54.08&38.95&29.35&73.92&47.41&46.57&41.59&73.87&54.01\\
&&&Focal&46.02&42.01&73.80&53.95&32.96&24.50&74.00&43.82&46.11&41.23&73.98&53.77\\
&&& Dice&48.01&47.28&71.22&55.55&48.59&48.48&71.22&56.10&48.07&47.79&71.29&55.72\\
&&& Weighted CE&46.07&42.18&73.82&54.02&40.54&30.07&73.98&48.20&46.31&41.68&73.95&53.98\\
&&& Weighted Focal&43.71&41.77&73.95&53.14&32.67&24.40&74.07&43.71&44.06&41.12&74.11&53.10\\
&&& Weighted Dice&47.97&47.22&71.02&55.40&48.69&48.34&71.02&56.02&48.04&47.70&71.10&55.61\\
&&& Fusion&48.12&47.25&71.93&55.77&46.34&44.25&71.86&54.15&48.19&47.89&72.38&56.15\\

\cmidrule(l){2-16} 
&\multirow{7}{*}{\textbf{\begin{tabular}[c]{@{}c@{}}SSP - VGG16\end{tabular}}}
&\multirow{7}{*}{{\begin{tabular}[c]{@{}c@{}}$\sim$24M\end{tabular}}} 
&CE&35.14&32.40&73.96&47.17&30.97&19.14&73.88&41.33&35.21&32.08&74.13&47.14\\
&&&Focal&34.56&32.22&74.02&46.93&29.11&19.42&73.91&40.81&34.67&31.79&74.22&46.89\\
&&&Dice&40.64&39.64&71.13&50.47&43.30&40.16&71.21&51.55&40.81&40.25&71.25&50.77\\
&&& Weighted CE&35.73&33.44&74.43&47.87&32.00&20.70&74.12&42.28&35.93&32.95&74.65&47.84\\
&&& Weighted Focal&34.90&34.35&74.36&47.87&29.18&20.72&74.06&41.32&35.11&33.80&74.56&47.82\\
&&& Weighted Dice&41.80&40.64&70.67&51.03&44.14&41.39&70.70&52.07&42.52&41.19&70.97&51.56\\
&&& Fusion&39.99&39.56&72.31&50.62&35.39&27.93&72.35&45.22&40.10&39.94&72.63&50.89\\

\cmidrule(l){2-16} 
&\multirow{7}{*}{\textbf{\begin{tabular}[c]{@{}c@{}}MFM (FFP + SSP) - VGG16\end{tabular}}}
&\multirow{7}{*}{{\begin{tabular}[c]{@{}c@{}}$\sim$24M\end{tabular}}} 
&CE&42.41&40.41&73.76&52.19&33.84&25.74&73.97&44.52&42.44&39.13&74.09&51.89\\
&&&Focal&41.61&39.11&73.77&51.50&30.85&23.13&74.04&42.67&41.64&37.74&74.13&51.17\\
&&&Dice&21.33&19.96&80.50&40.59&23.39&21.96&80.00&41.78&21.35&20.00&80.51&40.62\\
&&& Weighted CE&42.60&40.83&73.85&52.43&34.88&25.90&74.03&44.94&42.63&39.61&74.16&52.13\\
&&& Weighted Focal&42.22&40.83&73.85&52.30&30.79&23.74&74.10&42.88&42.25&39.51&74.18&51.98\\
&&& Weighted Dice&28.51&27.57&76.39&44.16&30.89&29.32&75.67&45.29&28.70&27.87&76.37&44.31\\
&&& Fusion&44.80&45.64&72.10&54.18&40.77&36.92&72.00&49.90&44.88&44.67&72.86&54.14\\

\cmidrule(l){2-16} 
&\multirow{7}{*}{\textbf{\begin{tabular}[c]{@{}c@{}}SSP - InceptionV3\end{tabular}}}
&\multirow{7}{*}{{\begin{tabular}[c]{@{}c@{}}$\sim$30M\end{tabular}}} 
&CE&49.54&42.13&74.10&55.26&42.80&33.23&74.17&50.07&49.56&41.24&74.34&55.05\\
&&&Focal&47.49&42.03&73.94&54.49&35.24&26.26&74.11&45.20&47.54&40.95&74.22&54.24\\
&&& Dice&51.22&49.71&71.39&57.44&51.87&\underline{\textbf{52.01}}&71.38&58.42&51.32&50.42&71.56&57.77\\
&&& Weighted CE&48.82&41.55&74.34&54.90&43.84&32.55&74.26&50.22&48.92&40.63&74.59&54.71\\
&&& Weighted Focal&45.06&40.39&74.19&53.21&34.27&24.89&74.29&44.48&45.27&39.25&74.52&53.01\\
&&& Weighted Dice&51.35&49.58&71.00&57.31&52.04&51.12&71.03&58.06&51.45&50.32&71.15&57.64\\
&&& Fusion&51.29&49.51&71.71&57.50&48.82&46.27&71.78&55.62&51.37&50.46&72.15&58.00\\

\cmidrule(l){2-16} 
&\multirow{7}{*}{\textbf{\begin{tabular}[c]{@{}c@{}}MFM (FFP + SSP) - InceptionV3\end{tabular}}}
&\multirow{7}{*}{{\begin{tabular}[c]{@{}c@{}}$\sim$30M\end{tabular}}} 
&CE&52.51&44.09&73.91&56.84&46.03&35.94&74.21&52.06&52.55&42.17&74.35&56.36\\
&&&Focal&51.56&43.74&73.89&56.40&35.89&27.15&74.41&45.82&51.63&41.74&74.43&55.93\\
&&& Dice&20.91&19.28&81.52&40.57&25.51&24.61&79.94&43.35&20.92&19.29&81.54&40.58\\
&&& Weighted CE&51.58&43.72&73.95&56.42&44.78&34.04&74.32&51.05&51.64&41.85&74.40&55.96\\
&&& Weighted Focal&51.01&43.62&73.89&56.18&35.76&26.60&74.22&45.53&51.08&41.60&74.35&55.68\\
&&& Weighted Dice&22.30&21.25&{80.73}&41.42&29.11&29.48&77.97&45.52&22.32&21.26&{{80.74}}&41.44\\
&&& Fusion&\underline{{53.22}}&51.25&71.84&58.77&50.05&48.83&72.03&56.97&\underline{\textbf{53.32}}&49.83&72.72&58.62\\

\cmidrule(l){2-16} 
&\multirow{7}{*}{\textbf{\begin{tabular}[c]{@{}c@{}}SSP - ResNet34\end{tabular}}}
&\multirow{7}{*}{{\begin{tabular}[c]{@{}c@{}}$\sim$24M\end{tabular}}} 
&CE&51.94&43.74&73.94&56.54&45.76&35.32&74.18&51.75&51.98&43.13&74.08&56.39\\
&&&Focal&50.99&43.45&73.93&56.12&35.76&27.78&74.33&45.96&50.99&42.34&74.19&55.84\\
&&& Dice&20.88&19.54&\underline{\textbf{82.08}}&40.83&22.00&20.83&\underline{81.89}&41.57&20.88&19.55&\underline{\textbf{82.08}}&40.83\\
&&& Weighted CE&51.45&43.27&74.03&56.25&45.76&35.01&74.13&51.63&51.43&42.55&74.20&56.06\\
&&& Weighted Focal&50.16&43.02&73.97&55.72&35.64&27.54&74.25&45.81&50.18&41.89&74.25&55.44\\
&&& Weighted Dice&52.04&50.75&70.97&57.92&52.35&{51.94}&71.00&58.43&52.09&51.25&71.07&58.14\\
&&& Fusion&52.58&50.69&71.77&58.35&50.19&48.91&71.94&57.01&52.60&\underline{{51.86}}&72.19&58.88\\

\cmidrule(l){2-16} 
&\multirow{7}{*}{\textbf{\begin{tabular}[c]{@{}c@{}} MFM (FFP + SSP) - ReNet34\end{tabular}}}
&\multirow{7}{*}{{\begin{tabular}[c]{@{}c@{}}$\sim$24M\end{tabular}}} 
&CE&52.40&44.02&73.91&56.78&47.52&36.40&74.12&52.68&52.43&42.12&74.34&56.30\\
&&&Focal&51.95&43.98&73.89&56.61&36.73&27.17&74.31&46.07&51.90&41.99&74.33&56.08\\
&&& Dice&29.68&29.43&77.06&45.39&31.62&25.39&75.15&44.05&29.84&29.63&77.07&45.52\\
&&& Weighted CE&52.63&44.12&73.94&56.90&47.60&37.25&74.12&52.99&52.56&42.22&74.36&56.38\\
&&& Weighted Focal&51.68&43.95&73.90&56.51&35.73&27.22&74.30&45.75&51.72&41.98&74.34&56.01\\
&&& Weighted Dice&51.44&50.60&71.96&58.00&\underline{52.97}&51.42&71.90&\underline{58.76}&51.54&49.37&72.75&57.89\\
&&& Fusion&{52.99}&\underline{{51.72}}& 71.69&\underline{58.80}&51.15&50.79&71.83&57.92&53.01&51.29&72.49&\underline{\textbf{58.93}}\\
\bottomrule
\end{tabular}%
}
    \captionof{table}{IoU performance (\%) on the WGM dataset~\cite{vafaie2022handwritten}. The maximum value of a column (i.e., class) is underlined. The overall maximum of a class with different post-processing is marked in bold and underlined. For example, the best mean IoU for the WGM-SYN dataset is for Fusion loss, with CRFH, and MFM-ResNet34 architecture at \underline{\textbf{58.93}}. The best performing configuration of prior work, marked with (\textsuperscript{*}), is 50.00.}
    \label{tab:res_wgm}
\end{table*}

\begin{table*}
      \resizebox{\linewidth}{!}{%
      \begin{tabular}{@{}cccllccc|lccc|lccc}
      \toprule
      &&  && \multicolumn{4}{c}{\textbf{IoU \%}}       &  \multicolumn{4}{c}{\textbf{With CRF (IoU \%)}} &  \multicolumn{3}{c}{\textbf{With CRFH (IoU \%)}}     \\  \cmidrule(lr){5-8} \cmidrule(l){9-12} \cmidrule(l){13-16} 
      Formulation                                                                          & Backbone & \# Parameters & Loss function  & PT & HT & BG& Mean &  PT & HT & BG & Mean  &  PT & HT & BG & Mean \\ \midrule
      \multirow{7}{*}{\textbf{\begin{tabular}[c]{@{}c@{}}3-Class\end{tabular}}}  
      & \multirow{7}{*}{\textbf{\begin{tabular}[c]{@{}c@{}}FCN-based~\cite{dutly2019phti,vafaie2022handwritten}\end{tabular}}}  & \multirow{6}{*}{{\begin{tabular}[c]{@{}c@{}}$\sim$295K\end{tabular}}} 
      & CE &62.56&88.09&98.40&83.02\textsuperscript{*}&52.68&89.68&99.26&80.46&62.72&90.58&99.05&84.11 \\
      &&& Focal &62.34&88.00&98.45&82.93&44.60&84.86&99.28&76.25&62.57&90.83&99.21& 84.21\\
      &&&Dice&60.45&87.29&97.85&81.86&61.24&88.83&98.17&82.74&60.52&87.88&97.97&82.12\\
      &&&Weighted CE&60.58&84.89&97.78&81.09&53.45&90.73&99.52&81.23&60.84&86.33&98.24&81.80\\
      &&&Weighted Focal&61.25&85.74&98.02&81.67&44.27&85.77&99.50&76.51&61.55&87.42&98.58&82.52\\
      &&&Weighted Dice &60.21&87.00&97.74&81.65&60.72&88.17&97.98&82.29&60.27&87.46&97.83&81.86\\
      &&&Fusion &61.52&88.39&98.38&82.76&58.94&92.55&99.37&83.62&61.67&90.45&98.91&83.68\\

      \midrule 
      \multirow{49}{*}{\textbf{\begin{tabular}[c]{@{}c@{}}\Large 4-Class\\(Ours)\end{tabular}}}  
      & \multirow{7}{*}{\textbf{\begin{tabular}[c]{@{}c@{}}FCN-based\end{tabular}}}  
      &\multirow{7}{*}{{\begin{tabular}[c]{@{}c@{}}$\sim$295K\end{tabular}}} 
      &CE &64.55&89.21&98.39&84.05&54.60&89.68&99.23&81.17&64.87&91.81&99.06&85.25\\
      &&&Focal &64.10&88.86&98.32&83.76&46.64&86.01&99.26&77.30&64.34&91.65&99.11&85.03\\
      &&&Dice &64.37&88.68&98.37&83.81&65.17&90.13&98.59&84.63&64.57&89.14&98.46&84.06\\
      &&&Weighted CE &63.78&87.52&98.16&83.15&54.77&90.50&99.42&81.56&64.12&89.43&98.71&84.09\\
      &&&Weighted Focal&63.68&87.71&98.19&83.20&48.18&87.10&99.45&78.24&64.05&89.72&98.80&84.19\\
      &&&Weighted Dice &64.21&88.57&98.30&83.69&65.08&90.10&98.53&84.57&64.44&89.06&98.39&83.96\\
      &&&Fusion &64.68&88.48&98.33&83.83&59.83&91.77&99.28&83.63&65.00&90.72&98.91&84.87\\

      \cmidrule(l){2-16} 
      &\multirow{7}{*}{\textbf{\begin{tabular}[c]{@{}c@{}}SSP - VGG16\end{tabular}}}
      &\multirow{7}{*}{{\begin{tabular}[c]{@{}c@{}}$\sim$24M\end{tabular}}} 
      &CE &47.36&81.02&98.23&75.54&32.20&81.80&99.52&71.18&47.44&84.20&99.06&76.90\\
      &&&Focal &47.17&81.46&98.18&75.60&28.51&79.66&99.53&69.23&47.21&8.458&99.05&76.95\\
      &&&Dice &52.83&80.12&97.17&76.71&54.13&84.14&97.95&78.74&52.99&81.45&97.49&77.31\\
      &&&Weighted CE &52.80&83.54&97.95&78.10&37.51&85.04&99.61&74.05&52.99&81.45&97.49&77.31\\
      &&&Weighted Focal &47.24&81.03&97.92&75.40&28.07&80.15&99.61&69.28&47.39&82.85&98.42&76.22\\
      &&&Weighted Dice &43.54&74.56&96.45&71.52&42.80&79.25&97.48&73.17&43.63&76.54&96.93&72.37\\
      &&&Fusion &49.36&80.54&97.82&75.91&38.66&84.94&99.59&74.39&49.52&83.31&98.57&77.14\\

      \cmidrule(l){2-16} 
      &\multirow{7}{*}{\textbf{\begin{tabular}[c]{@{}c@{}}MFM (FFP + SSP) - VGG16\end{tabular}}}
      &\multirow{7}{*}{{\begin{tabular}[c]{@{}c@{}}$\sim$24M\end{tabular}}} 
      &CE &61.52&88.90&98.67&83.03&45.72&87.01&99.54&77.42&61.69&91.27&99.34&84.10\\
      &&&Focal &60.99&88.45&98.65&82.69&37.41&82.76&99.55&73.24&61.08&90.85&99.34&83.75\\
      &&& Dice &59.73&88.43&98.65&82.27&61.66&90.50&98.94&83.70&60.42&90.75&99.30&83.49\\
      &&&Weighted CE &57.15&87.28&98.52&80.98&41.46&86.51&99.61&75.86&57.54&88.94&98.99&81.83\\
      &&&Weighted Focal &57.90&87.35&98.49&81.24&36.60&82.71&99.60&72.97&58.35&89.05&98.97&82.12\\
      &&&Weighted Dice &58.14&87.67&98.62&81.48&60.19&90.31&98.98&83.16&58.82&90.36&99.37&82.85\\
      &&&Fusion &59.80&88.33&98.62&82.25&51.14&90.13&99.59&80.28&60.46&90.25&99.18&83.30\\

      \cmidrule(l){2-16} 
      &\multirow{7}{*}{\textbf{\begin{tabular}[c]{@{}c@{}}SSP - InceptionV3\end{tabular}}}
      &\multirow{7}{*}{{\begin{tabular}[c]{@{}c@{}}$\sim$30M\end{tabular}}} 
      &CE &72.82&92.44&98.73&87.99&63.32&92.91&99.55&85.26&72.77&94.90&99.29&88.99\\
      &&&Focal &72.20&92.14&98.69&87.68&56.95&90.34&99.54&82.28&72.18&94.86&99.32&88.79\\
      &&&Dice &71.11&91.11&98.55&86.92&71.54&93.53&99.07&88.05&71.11&92.22&98.79&87.37\\
      &&&Weighted CE &70.32&90.53&98.34&86.39&61.39&92.52&\underline{\textbf{99.63}}&84.52&70.35&92.15&98.72&87.07\\
      &&&Weighted Focal &71.64&91.06&98.39&87.03&56.10&90.48&99.60&82.06&71.62&92.68&98.79&87.69\\
      &&&Weighted Dice &71.96&91.60&98.59&87.39&72.15&93.29&98.96&88.14&71.92&92.35&98.76&87.68\\
      &&&Fusion &71.19&91.10&98.51&86.93&65.80&93.97&99.59&86.45&71.18&93.32&99.01&87.84\\

      \cmidrule(l){2-16} 
      &\multirow{7}{*}{\textbf{\begin{tabular}[c]{@{}c@{}} MFM (FFP + SSP) - InceptionV3\end{tabular}}}
      &\multirow{7}{*}{{\begin{tabular}[c]{@{}c@{}}$\sim$30M\end{tabular}}} 
      &CE &73.10&\underline{92.66}&98.77&\underline{88.18}&63.48&92.89&99.55&85.31&73.05&94.89&99.36&\underline{\textbf{89.10}}\\
      &&&Focal &72.80&92.50&98.75&88.01&57.47&90.58&99.54&82.53&72.77&94.74&99.35&88.95\\
      &&&Dice &72.56&92.20&98.70&87.82&72.38&93.60&99.00&88.33&72.52&94.73&99.37&88.87\\
      &&&Weighted CE &72.66&91.90&98.54&87.70&62.65&93.04&99.62&85.10&72.63&93.31&98.94&88.29\\
      &&&Weighted Focal &72.55&92.13&98.62&87.77&59.35&91.64&99.59&83.53&72.50&93.77&99.07&88.45\\
      &&&Weighted Dice &72.59&92.31&98.70&87.87&72.60&94.00&99.07&\underline{88.56}&72.54&94.60&99.32&88.82\\
      &&&Fusion &72.55&92.25&98.71&87.83&65.57&93.49&99.53&86.19&72.49&94.62&99.35&88.82\\

      \cmidrule(l){2-16} 
      &\multirow{7}{*}{\textbf{\begin{tabular}[c]{@{}c@{}}SSP - ResNet34\end{tabular}}}
      &\multirow{7}{*}{{\begin{tabular}[c]{@{}c@{}}$\sim$24M\end{tabular}}} 
      &CE &73.02&92.27&98.71&88.00&63.68&92.57&99.55&85.27&72.97&94.71&99.27&88.98\\
      &&&Focal &72.74&92.26&98.71&87.90&58.54&90.95&99.55&83.01&72.71&94.88&99.32&88.97\\
      &&&Dice &72.40&91.91&98.65&87.65&72.62&93.38&98.98&88.33&72.35&92.55&98.79&87.90\\
      &&&Weighted CE &72.43&91.34&98.45&87.40&62.94&92.97&\underline{\textbf{99.63}}&85.18&72.43&92.82&98.80&88.02\\
      &&&Weighted Focal &72.39&91.38&98.43&87.40&58.85&91.38&99.61&83.28&72.35&92.94&98.81&88.03\\
      &&&Weighted Dice &71.86&91.89&98.65&87.47&71.99&93.32&98.96&88.09&71.79&92.56&98.80&87.72\\
      &&&Fusion &72.96&91.88&98.60&87.81&68.64&\underline{94.86}&99.58&87.69&72.92&94.08&99.10&88.70\\

      \cmidrule(l){2-16} 
      &\multirow{7}{*}{\textbf{\begin{tabular}[c]{@{}c@{}}MFM (FFP + SSP) - ResNet34\end{tabular}}}
      &\multirow{7}{*}{{\begin{tabular}[c]{@{}c@{}}$\sim$24M\end{tabular}}} 
      &CE &72.81&92.56&\underline{98.78}&88.05&63.04&92.94&99.55&85.17&72.75&\underline{\textbf{94.93}}&\underline{{99.39}}&89.02\\
      &&&Focal &73.04&92.46&98.75&88.08&53.02&89.04&99.55&80.54&73.00&94.75&99.35&89.03\\
      &&&Dice &72.96&92.38&98.72&88.02&\underline{73.16}&93.35&98.93&88.48&72.93&94.79&99.36&89.03\\
      &&&Weighted CE &72.96&91.96&98.69&87.87&64.49&92.47&99.56&85.51&72.90&94.19&99.27&88.79\\
      &&&Weighted Focal &73.18&92.10&98.63&87.97&54.96&89.71&99.60&81.42&73.16&93.72&99.06&88.65\\
      &&&Weighted Dice &72.78&92.32&98.71&87.94&72.95&93.66&99.00&88.53&72.75&94.67&99.34&88.92\\
      &&&Fusion &\underline{\textbf{73.26}}&92.45&98.73&{88.15}&68.38&94.85&99.56&87.60&\underline{73.21}&94.59&99.31&{{89.04}}\\
                                                                         
      \bottomrule
      \end{tabular}%
      }
          \captionof{table}{IoU performance (\%) on the SignaTR6K dataset. Partial results for this dataset were presented in the main text and this table provides the full results. The maximum value of a column (i.e., class) is underlined, and the overall maximum of a class with different post-processing is denoted in bold and underlined. For example, the best mean IoU for the SignaTR6K dataset is for CE loss, with CRFH, and MFM-InceptionV3 architecture at \underline{\textbf{89.10}}. The best performing configuration of prior work, i.e., excluding results for Fusion loss and CRFH, marked with (\textsuperscript{*}), is 83.02.}
          \label{tab2:signaTR6K}
      \end{table*}

\begin{figure*}
\captionsetup[subfigure]{aboveskip=-1pt,belowskip=-1pt}
  \centering
  \begin{subfigure}{.2\textwidth}
    \centering
    \fbox{\includegraphics[width=\linewidth]{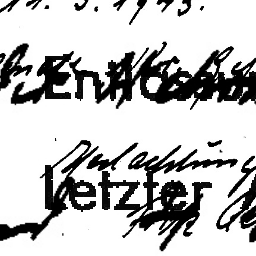}} 
    \caption{}
    \label{fig:sub1}
  \end{subfigure}
  \begin{subfigure}{.2\textwidth}
    \centering
    \fbox{\includegraphics[width=\linewidth]{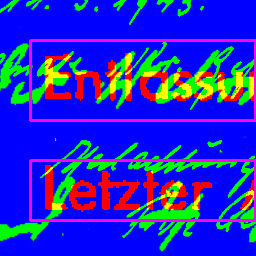}} 
    \caption{}
    \label{wgm_an:b}
  \end{subfigure}
  \begin{subfigure}{.2\textwidth}
    \centering
    \fbox{\includegraphics[width=\linewidth]{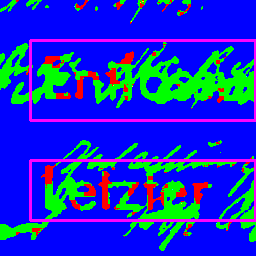}} 
    \caption{}
    \label{wgm:c}
  \end{subfigure}
  \begin{subfigure}{.2\textwidth}
    \centering
    \fbox{\includegraphics[width=\linewidth]{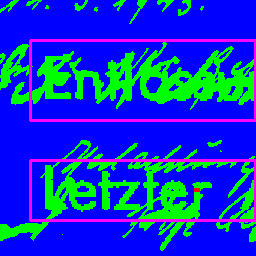}} 
    \caption{}
    \label{wgm:d}
  \end{subfigure}
  
  \begin{subfigure}{.2\textwidth}
    \centering
    \fbox{\includegraphics[width=\linewidth]{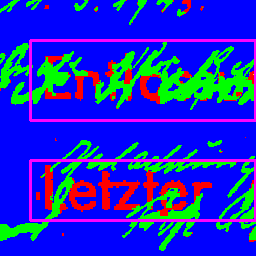}} 
    \caption{}
    \label{fig:sub1}
  \end{subfigure}
  \begin{subfigure}{.2\textwidth}
    \centering
    \fbox{\includegraphics[width=\linewidth]{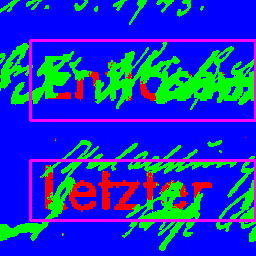}} 
    \caption{}
    \label{fig:sub1}
  \end{subfigure}
  \begin{subfigure}{.2\textwidth}
    \centering
    \fbox{\includegraphics[width=\linewidth]{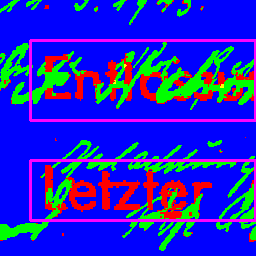}} 
    \caption{}
    \label{fig:sub1}
  \end{subfigure}
  \begin{subfigure}{.2\textwidth}
    \centering
    \fbox{\includegraphics[width=\linewidth]{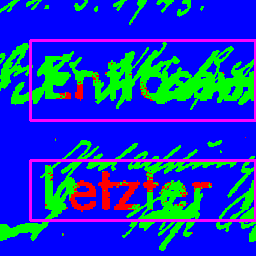}} 
    \caption{}
    \label{fig:sub1}
  \end{subfigure}
  
  \begin{subfigure}{.2\textwidth}
    \centering
    \fbox{\includegraphics[width=\linewidth]{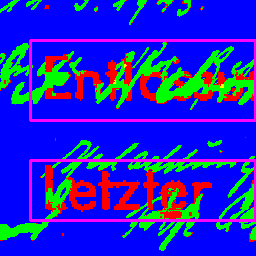}} 
    \caption{}
    \label{fig:sub1}
  \end{subfigure}
  \begin{subfigure}{.2\textwidth}
    \centering
    \fbox{\includegraphics[width=\linewidth]{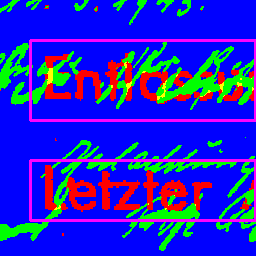}} 
    \caption{}
    \label{fig:sub1}
  \end{subfigure}
  \begin{subfigure}{.2\textwidth}
    \centering
    \fbox{\includegraphics[width=\linewidth]{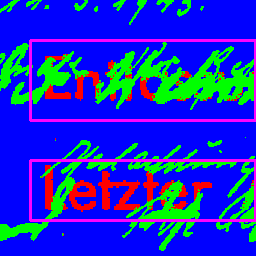}} 
    \caption{}
    \label{fig:sub1}
  \end{subfigure}
  \begin{subfigure}{.2\textwidth}
    \centering
    \fbox{\includegraphics[width=\linewidth]{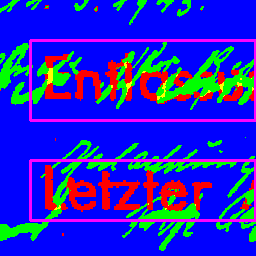}} 
    \caption{}
    \label{fig:sub1}
  \end{subfigure}
  \begin{subfigure}{.2\textwidth}
    \centering
    \fbox{\includegraphics[width=\linewidth]{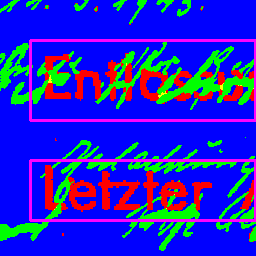}} 
    \caption{}
    \label{fig:sub1}
  \end{subfigure}
  \begin{subfigure}{.2\textwidth}
    \centering
    \fbox{\includegraphics[width=\linewidth]{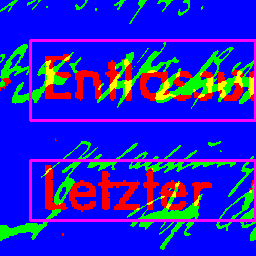}} 
    \caption{}
    \label{fig:sub1}
  \end{subfigure}
  \begin{subfigure}{.2\textwidth}
    \centering
    \fbox{\includegraphics[width=\linewidth]{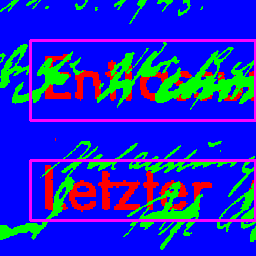}}
    \caption{}
    \label{fig:sub1}
  \end{subfigure}
  \begin{subfigure}{.2\textwidth}
    \centering
    \fbox{\includegraphics[width=\linewidth]{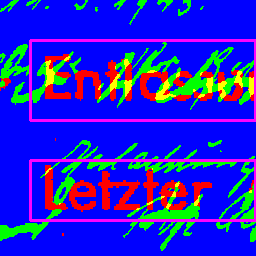}}
    \caption{}
    \label{wgm:p}
  \end{subfigure}
  
  \caption{Example visual comparisons on the test set of the WGM-SYN dataset for our approach compared to the ground truth and prior works. (a) Input image; (b) Ground truth; (c) \& (d) 3-class FCN-based~\cite{dutly2019phti} with CE loss without (c) and with (d) CRF post-processing; (e) \& (f) 3-class FCN-based~\cite{vafaie2022handwritten} with CE loss without (e) and with (f) CRF post-processing; (g), (h), \& (i) Our FCN-based 4-class formulation with CE loss without CRF (g), with CRF (h), and with CRFH (i); (j), (k), \& (l) SSP-ResNet34 with CE loss without CRF (j), with CRF (k), and with CRFH (l); (m) MFM-ResNet34 with CE loss without CRF; (n), (o), \& (p) MFM-ResNet34 with Fusion loss without CRF (n), with CRF (o), and with CRFH (p).}
  
\label{fig:wgm_an}
\end{figure*}

\begin{figure*}
\captionsetup[subfigure]{aboveskip=-1pt,belowskip=-1pt}
\centering
\begin{subfigure}{.20\textwidth}
  \centering
  \fbox{\includegraphics[width=\linewidth]{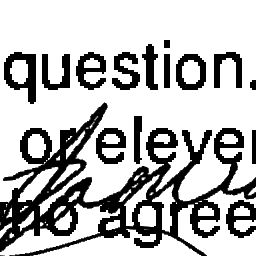}}
  \caption{}
  \label{fig:sub1}
\end{subfigure}
\begin{subfigure}{.20\textwidth}
  \centering
  \fbox{\includegraphics[width=\linewidth]{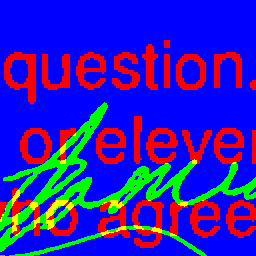}} 
  \caption{}
  \label{tr2:b}
\end{subfigure}
\begin{subfigure}{.20\textwidth}
  \centering
  \fbox{\includegraphics[width=\linewidth]{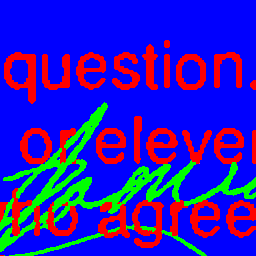}}
  \caption{}
  \label{fig:sub2}
\end{subfigure}
\begin{subfigure}{.20\textwidth}
  \centering
  \fbox{\includegraphics[width=\linewidth]{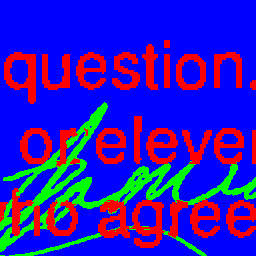}} 
  \caption{}
  \label{fig:sub2}
\end{subfigure}

\begin{subfigure}{.20\textwidth}
  \centering
  \fbox{\includegraphics[width=\linewidth]{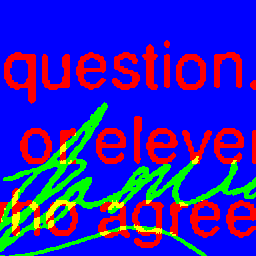}}
  \caption{}
  \label{fig:sub2}
\end{subfigure}
\begin{subfigure}{.2\textwidth}
  \centering
  \fbox{\includegraphics[width=\linewidth]{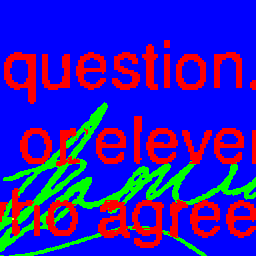}} 
  \caption{}
  \label{fig:sub2}
\end{subfigure}
\begin{subfigure}{.2\textwidth}
  \centering
  \fbox{\includegraphics[width=\linewidth]{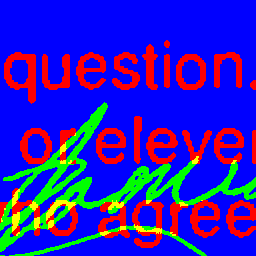}} 
  \caption{}
  \label{fig:sub2}
\end{subfigure}
\begin{subfigure}{.2\textwidth}
  \centering
  \fbox{\includegraphics[width=\linewidth]{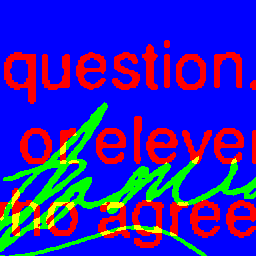}} 
  \caption{}
  \label{fig:sub2}
\end{subfigure}

\begin{subfigure}{.2\textwidth}
  \centering
  \fbox{\includegraphics[width=\linewidth]{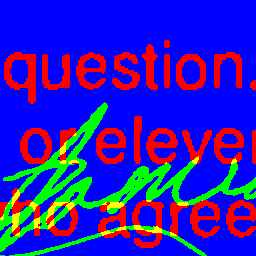}} 
  \caption{}
  \label{fig:sub2}
\end{subfigure}
\begin{subfigure}{.2\textwidth}
  \centering
  \fbox{\includegraphics[width=\linewidth]{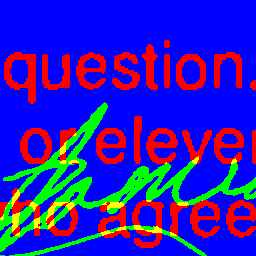}} 
  \caption{}
  \label{fig:sub2}
\end{subfigure}
\begin{subfigure}{.2\textwidth}
  \centering
  \fbox{\includegraphics[width=\linewidth]{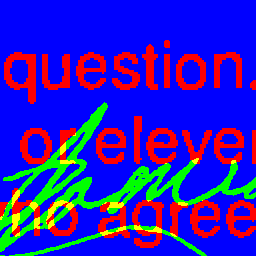}} 
  \caption{}
  \label{fig:sub2}
\end{subfigure}
\begin{subfigure}{.2\textwidth}
  \centering
  \fbox{\includegraphics[width=\linewidth]{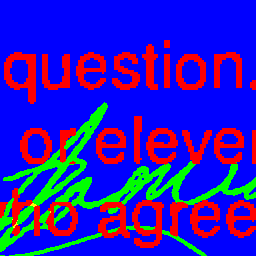}} 
  \caption{}
  \label{fig:sub2}
\end{subfigure}

\begin{subfigure}{.2\textwidth}
  \centering
  \fbox{\includegraphics[width=\linewidth]{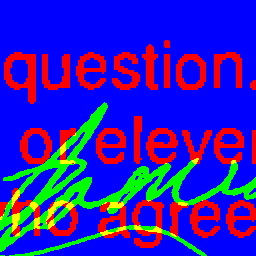}} 
  \caption{}
  \label{fig:sub2}
\end{subfigure}
\begin{subfigure}{.2\textwidth}
  \centering
  \fbox{\includegraphics[width=\linewidth]{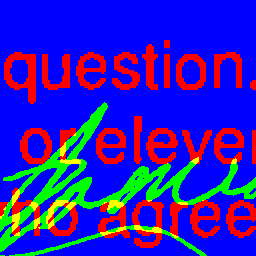}} 
  \caption{}
  \label{fig:sub2}
\end{subfigure}
\begin{subfigure}{.2\textwidth}
  \centering
  \fbox{\includegraphics[width=\linewidth]{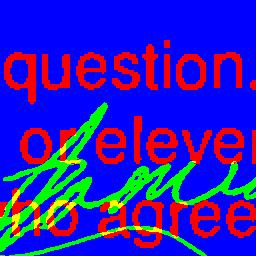}} 
  \caption{}
  \label{fig:sub2}
\end{subfigure}
\begin{subfigure}{.2\textwidth}
  \centering
  \fbox{\includegraphics[width=\linewidth]{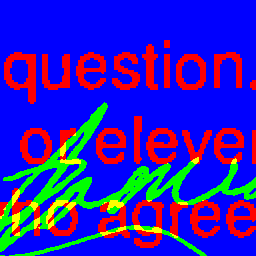}}
  \caption{}
  \label{fig:sub2}
\end{subfigure} 

\caption{Additional visual results on the test set of the SignaTR200 dataset for our approach compared to the ground truth and prior works.  (a) Input image; (b) Ground truth; (c) \& (d) 3-class FCN-based~\cite{dutly2019phti,vafaie2022handwritten} with CE loss without (c) and with (d) CRF post-processing; (e), (f), \& (g) Our FCN-based 4-class formulation with CE loss without CRF (e), with CRF (f), and with CRFH (g); (h), (i), \& (j) SSP-ResNet34 with CE loss without CRF (h), with CRF (i), and with CRFH (j); (k), (l), \& (m) MFM-ResNet34 with CE loss without CRF (k), with CRF (l), and with CRFH (m); (n), (o), \& (p) MFM-ResNet34 with Fusion loss without CRF (n), with CRF (o), and with CRFH (p).}
\label{fig:tr200_2}

\end{figure*}

\subsection{Additional Results}
Tables~\ref{tab:res_wgm} and~\ref{tab2:signaTR6K} show the IoU values for WMG-SYN and SignaTR6K datasets, respectively. 
Overall trends of the results show that going from the three-class formulation to the four-class formulation improves the IoU values. 
Additionally, using larger model backbones generally improves the segmentation performance. 
Among all the model architectures, the ResNet34 and InceptionV3 backbones achieve the highest performance, which we attribute to their residual connections and varied-size convolutions as they can better incorporate the finer features from the image.

\subsection{Visual Comparisons}
Figures~\ref{fig:wgm_an} and~\ref{fig:tr200_2} provide additional visual comparisons for WGM-SYN and SignaTR6K datasets. 
Figures~\ref{wgm_an:b} and~\ref{tr2:b} show the ground truth, and we can visually observe a trend that the performance on the HT and PT overlapping regions improves from (c) to (p). 
It is also visually noticeable that CRF post-processing aggressively relabels pixels, and CRFH generally improves the results. 

\subsection{Dataset Availability}
The SignaTR6K dataset is available for download through this link: \url{https://forms.office.com/r/2a5RDg7cAY}. 

\end{document}